\documentclass[lettersize,journal]{IEEEtran}
\usepackage{amsmath,amsfonts}
\usepackage{algorithmic}
\usepackage{array}
\usepackage[caption=false,font=normalsize,labelfont=sf,textfont=sf]{subfig}
\usepackage{textcomp}
\usepackage{stfloats}
\usepackage{url}
\usepackage{verbatim}
\usepackage{graphicx}
\def\BibTeX{{\rm B\kern-.05em{\sc i\kern-.025em b}\kern-.08em
    T\kern-.1667em\lower.7ex\hbox{E}\kern-.125emX}}
\usepackage{balance}

\usepackage{comment}
\usepackage{amssymb} % define this before the line numbering.
\usepackage{color}
\usepackage{booktabs}

\usepackage{multirow}
\usepackage{wrapfig}
\usepackage{makecell}
\usepackage{bbding}

\usepackage{adjustbox}
\usepackage{array}
\usepackage{xcolor,colortbl}
\usepackage{balance}

\newcommand{\Dmat}{{\bf D}}

\newcommand{\Gmat}[0]{{{\bf G}}}

\newcommand{\Kmat}[0]{{{\bf K}}}

\newcommand{\Mmat}[0]{{{\bf M}}}

\newcommand{\Qmat}[0]{{{\bf Q}}}

\newcommand{\Vmat}[0]{{{\bf V}}}
\newcommand{\Wmat}[0]{{{\bf W}}}
\newcommand{\Xmat}{{\bf X}}
\newcommand{\Ymat}[0]{{{\bf Y}}}
\newcommand{\Zmat}{{\bf Z}}

\newcommand{\gv}[0]{{\boldsymbol{g}}}

\newcommand{\xv}{\boldsymbol{x}}
\newcommand{\yv}{\boldsymbol{y}}

\newcommand{\Phimat}{\boldsymbol{\Phi}}

\begin{document}
\title{Sparse Transformer for Ultra-sparse Sampled \\Video Compressive Sensing}

\author{Miao Cao, Siming Zheng, Lishun Wang, Ziyang Chen, \\David Brady,~\IEEEmembership{Fellow,~IEEE} and Xin Yuan,~\IEEEmembership{Senior~Member,~IEEE}% <-this % stops a space
\thanks{*This work was supported by National Key R\&D Program of China (2024YFF0505603), the National Natural Science Foundation of China (grant number 62271414), Zhejiang Provincial Distinguished Young Scientist Foundation (grant number LR23F010001), Zhejiang `Pioneer' and `Leading Goose' R\&D Program (grant number 2024SDXHDX0006, 2024C03182), the Key Project of Westlake Institute for Optoelectronics (grant number 2023GD007), the 2023 International Sci-tech Cooperation Projects under the purview of the `Innovation Yongjiang 2035' Key R\&D Program (grant number 2024Z126).}% <-this % stops a space
\thanks{$^{1}$Miao Cao is with Zhejiang University, Hangzhou 310058, China and Westlake University, Hangzhou 310030, China. (e-mail: caomiao@westlake.edu.cn)}%
\thanks{$^{2}$Siming Zheng is with vivo Mobile Communication Co., Ltd, Hangzhou 310000, China. (e-mail: zhengsiming@vivo.com)}%
\thanks{$^{3}$Lishun Wang, Ziyang Chen, and Xin Yuan are with School of Engineering and Research Center for Industries of the Future, Westlake University, Hangzhou 310030, China. (e-mail: wanglishun@westlake.edu.cn, czy2506855759@foxmail.com, and xyuan@westlake.edu.cn)}%
\thanks{$^{4}$David Brady is with Wyant College of Optical Sciences, University of Arizona, Tucson, AZ 85721, USA. (e-mail: djbrady@arizona.edu)}%
\thanks{$^{5}$Corresponding author: Xin Yuan.}%
}

\markboth{IEEE TRANSACTIONS ON MULTIMEDIA}%
{How to Use the IEEEtran \LaTeX \ Templates}

\maketitle

\begin{abstract}
Digital cameras consume $\sim 0.1$ microjoule per pixel to capture and encode video, resulting in a power usage of $\sim 20$W for a 4K sensor operating at 30 fps. Imagining gigapixel cameras operating at 100-1000 fps, the current processing model is unsustainable. To address this, physical layer compressive measurement has been proposed to reduce power consumption per pixel by 10-100$\times$. Video Snapshot Compressive Imaging (SCI) introduces high frequency modulation in the optical sensor layer to increase effective frame rate. A commonly used sampling strategy of video SCI is Random Sampling (RS) where each mask element value is randomly set to be 0 or 1. Similarly, image inpainting (I2P) has demonstrated that images can be recovered from a fraction of the image pixels. Inspired by I2P, we propose Ultra-Sparse Sampling (USS) regime, where at each spatial location, only one sub-frame is set to 1 and all others are set to 0. We then build a Digital Micro-mirror Device (DMD) encoding system to verify the effectiveness of our USS strategy. Ideally, we can decompose the USS measurement into sub-measurements for which we can utilize I2P algorithms to recover high-speed frames. However, due to the mismatch between the DMD and CCD, the USS measurement cannot be perfectly decomposed. To this end, we propose {\em BSTFormer}, a sparse TransFormer that utilizes local Block attention, global Sparse attention, and global Temporal attention to exploit the sparsity of the USS measurement. Extensive results on both simulated and real-world data show that our method significantly outperforms all previous state-of-the-art algorithms. Additionally, an essential advantage of the USS strategy is its higher dynamic range than that of the RS strategy. Finally, from the application perspective, the USS strategy is a good choice to implement a complete video SCI system on chip due to its fixed exposure time. Code is available at https://github.com/mcao92/BSTFormer.
\end{abstract}


\begin{figure}[!htbp]
    \centering
    \includegraphics[width=\linewidth] {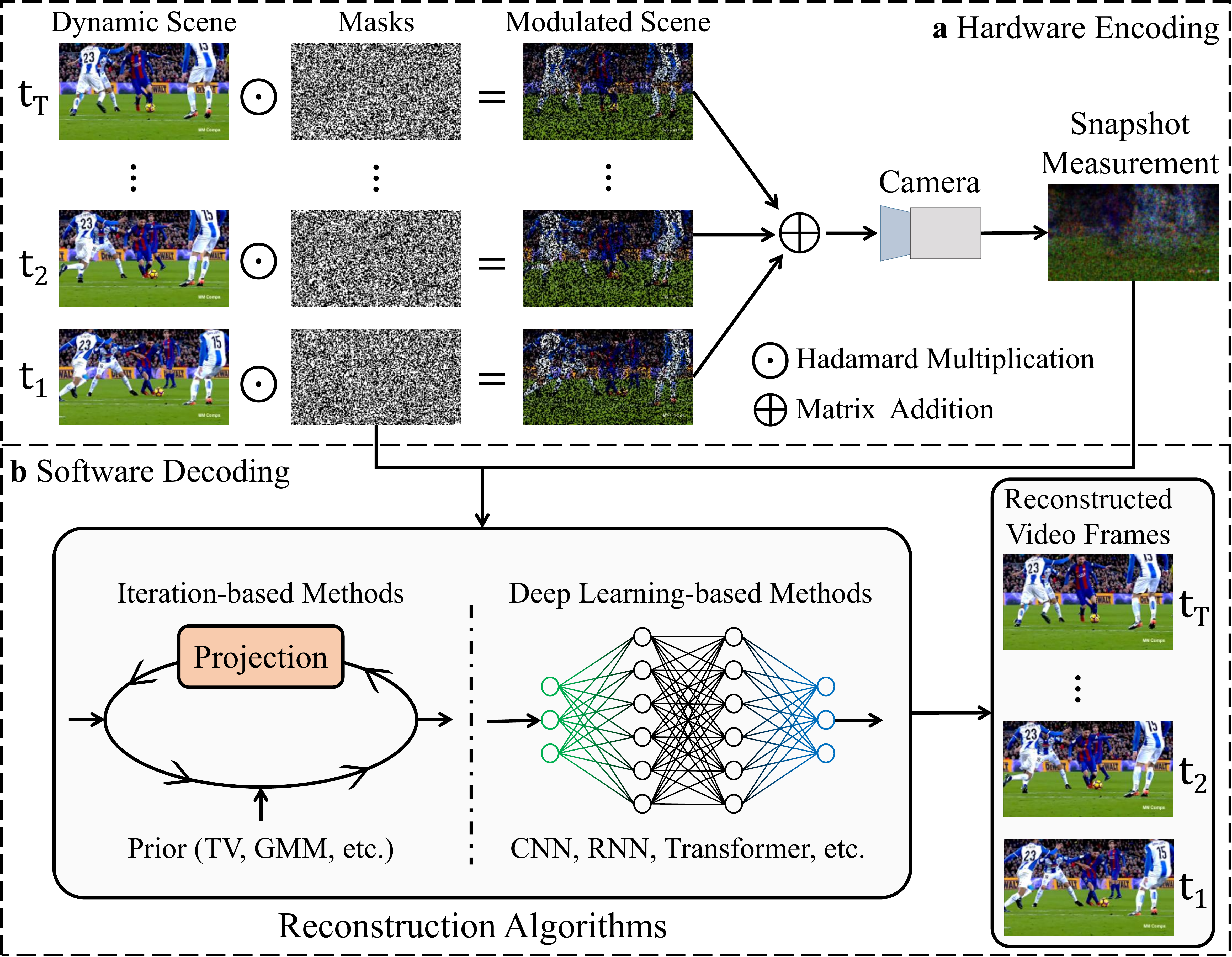}
    \caption{Principle of the video SCI system: a) In the encoding process, a dynamic scene occurring within a single exposure time (from $t_1$ to $t_T$) is first modulated by different masks. The modulated scene is then integrated over this time period. Finally, a low-speed camera captures the integrated scene as a snapshot compressed measurement. b) In the decoding process, the captured snapshot compressed measurement and the corresponding modulation masks are fed into a reconstruction algorithm to obtain the desired video frames.}
    \label{fig:sci}
\end{figure}

\section{Introduction\label{Sec:Intro}}

\IEEEPARstart{D}{ata} bandwidth and electrical power are significant barriers to achieving high-speed, high-resolution imaging. To address this, physical layer compression and computational imaging have been proposed~\cite{ashtiani2022chip}.
A key challenge in computational imaging arises from the mismatch between 2D sensor arrays and the multidimensional data volumes of objects. In this work, we focus on video Snapshot Compressive Imaging (SCI)~\cite{yuan2021snapshot}, which uses a low-speed camera to capture high-speed scenes. As illustrated in Fig.~\ref{fig:sci}, there are two steps in the video SCI system: hardware encoding and software decoding. In the hardware encoding process, the dynamic scene (occurring within a single exposure time, from $t_1$ to $t_T$) is modulated by random binary masks. These masks are composed of 0s (blocking light) and 1s (transmitting light), with Random Sampling (RS) commonly used for modulation. The modulated scene is then integrated over this period. Finally, a low-speed camera captures the integrated scene as a snapshot compressed measurement. In the software decoding stage, the captured snapshot measurement, along with the corresponding modulation masks, are fed into a reconstruction algorithm to recover the desired video frames.

While one goal of video SCI is to capture high-speed effects, another important goal is to {\em reduce bandwidth and data throughput}. To achieve this, coding strategies include the following methods:
\begin{itemize}
    \item[a)] Sampling low spatial-resolution high-speed scenes and applying \textit{image/video super-resolution algorithms} to retrieve high spatial-temporal scenes. A challenge here is that we still need a high-speed camera.
    \item[b)] Different from case a), randomly sub-sampling a high spatial-resolution image at a high speed. Afterward, \textit{image inpainting (I2P) methods}~\cite{quan2024deep} are used to recover high spatial-temporal scenes.
    \item[c)] Both of the above cases require a high-speed camera. Inspired by I2P, we propose an Ultra-Sparse Sampling (USS) strategy for video SCI. Specifically, we can use a low-speed camera to capture a ``measurement'', which is a combined sub-image at different time slots. By setting the missing pixels in each sub-frame differently, an {\em ultra-sparse mask set} can be employed, where at each spatial location, only one sub-frame is set to 1 and all others are set to 0.
    Ideally, we can easily decompose the measurement into sub-measurements for which we can utilize I2P algorithms to recover high-speed frames. This is also the main goal of video SCI.
\end{itemize}

Video SCI uses code division multiplexing to capture multiple temporal frames in a snapshot. {\em Dynamic range} is a critical issue in such multiplexing systems. 
Assuming $T$ frames normalized into $[0,1]$ are modulated and collapsed into a single SCI measurement, the dynamic range can be $[0,T]$. However, the measurement can only represent the range $[0,1]$. Some pixels will be over-exposed and lead to degraded estimation. In contrast, if the USS strategy is used, the dynamic range of the measurement will still be $[0,1]$. While intuitively, we may think that USS strategy is better, there are some cases, such as dark or low-light environment where we do need a longer exposure time. Therefore, it will be the applications that determine which kind of sampling strategy should we use. This paper explores dynamic range performance using experimental data captured by our video SCI system.

For practical implementation of video SCI, one must {\em miniaturize the imaging system}, ideally on a chip.  One challenge to hinder this is the modulation device, such as digital micro-mirror device (DMD), commonly used in previous systems. The principle of video SCI is to {\bf randomly access each pixel at high speed and also to randomly set the exposure time for each pixel}. Therefore, this modulation device is not really necessary if we can access each pixel in a high speed. However, random exposure time is difficult. Towards this end, the USS strategy provides an effective solution to implement video SCI on a chip.

This paper aims to compare the above-mentioned two sampling strategies (RS and USS) in video SCI applications. More importantly, it seeks to bridge the gap between these strategies and explore how to combine them to advance the high-throughput imaging techniques. Another motivation of this paper is to move one step further to miniaturize video SCI systems. The specific contributions of this paper can be summarized as follows:
\begin{itemize}
    \item We introduce I2P into video SCI via specially designing the modulation patterns. Specifically, at each spatial location, only one sub-frame is set to 1, and all others are set to 0.
    \item Due to the mismatch between the DMD and CCD in real-built video SCI systems, the USS measurements cannot be ideally decomposed. To address this, we propose a sparse Trans{\bf Former} architecture ({\bf BSTFormer}), which includes local {\bf B}lock attention, global {\bf S}parse attention, and global {\bf T}emporal attention to better exploit the sparsity in USS measurements.
    \item We release a real-world testing dataset including the USS data with large spatial size ($800\times800$) and different compression ratios (\{10, 20, 30, 40, 50\}) to promote research on mask optimization in video SCI.
    \item Extensive results on both simulated and real-world data demonstrate that our method outperforms all previous state-of-the-art (SOTA) algorithms. Furthermore, the experimental results show that our proposed USS masking strategy achieves a higher dynamic range compared to RS one.
\end{itemize}

\section{Related Work}
\label{Sec:related}

\subsection{Video Snapshot Compressive Imaging}
\label{Sec:RS}
A variety of video SCI systems with different mask designs have been developed over the years. For example, Reddy et al. were the first to use Spatial Light Modulators to provide spatio-temporal modulation with random masks~\cite{reddy2011p2c2}. Similarly, Hitomi et al. employed normalized masks in a liquid crystal on silicon (LCoS) device~\cite{hitomi2011video}. Llull et al. modulated the light path by shifting a random mask on the translation stage~\cite{llull2013coded}. In the same way, Koller et al. used a normalized mask on a translation stage to modulate the high-speed scenes~\cite{koller2015high}. 
In contrast to these approaches, and inspired by I2P, we propose an ultra-sparse masking strategy. This strategy provides better dynamic range performance and easier implementation on chip in the future.

\begin{figure}[!htbp]
    \centering
    \includegraphics[width=\linewidth] {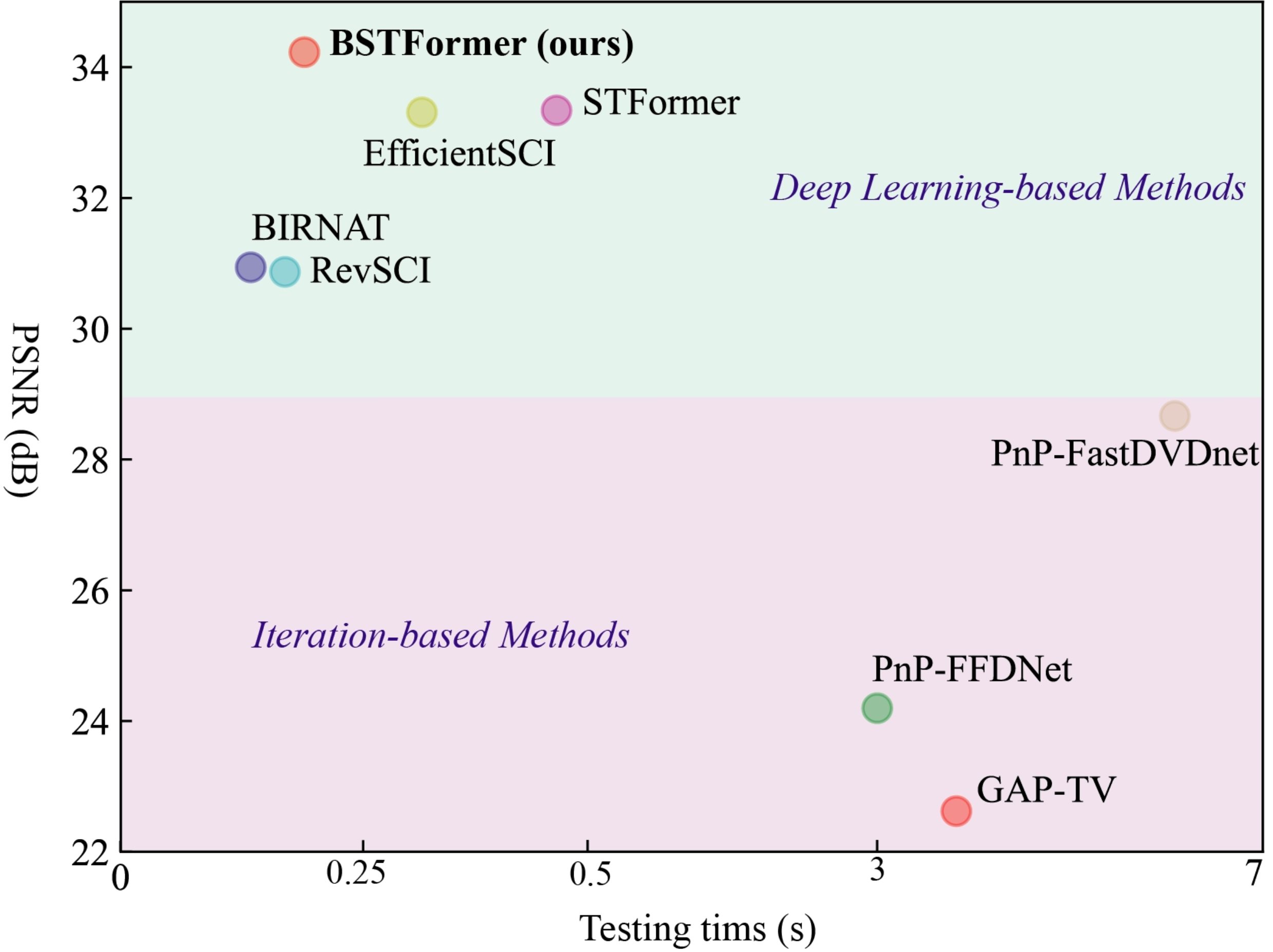}
    \caption{Comparison of reconstruction quality (average PSNR in dB on six simulated testing datasets) and inference time of several SOTA deep learning-based video SCI reconstruction algorithms. Our proposed BSTFormer can achieve higher reconstruction quality with shorter inference time.}
    \label{fig:psnr_time-1}
\end{figure}

In the decoding process, traditional model-based reconstruction methods utilize numerous regularization items, including total variation~\cite{Yuan16ICIP_GAP}, Gaussian mixture model~\cite{yang2014compressive}, and non-local low rank~\cite{liu2018rank}, etc. Although these model-based methods have good flexibility and can reconstruct videos with different resolutions and compression ratios (Crs), they usually require long reconstruction time and can only achieve poor or middle reconstruction quality. The plug-and-play (PnP) frameworks, such as PnP-FFDnet~\cite{yuan2020plug}, PnP-FastDVDnet~\cite{yuan2021plug}, and online PnP~\cite{wu2023adaptive} make the model more flexible by integrating a pre-trained deep image denoiser into an iterative optimization algorithm. Nevertheless, PnP-based methods are still time consuming due to their iterative nature. To overcome these drawbacks, deep learning-based methods~\cite{cao2024towards,cheng2020birnat,wang2023efficientsci,cao2025simple} have been proposed to solve the reconstruction problem. For example, Qiao et al. proposed the first end-to-end (E2E) neural network which utilizes a simple U-shaped structure for fast reconstruction~\cite{qiao2020deep}. Cheng et al. designed a novel bidirectional recurrent neural network for frame-by-frame reconstruction~\cite{cheng2020birnat} and later developed a 3D CNN-based approach with memory-efficient reversible networks~\cite{cheng2021memory}. Wang et al. proposed MetaSCI with light-weight meta-modulation parameters that can quickly adapt to new masks and be ready to scale to large data~\cite{wang2021metasci}.
More recently, Wang et al. introduced the first Transformer-based reconstruction network with space-time factorization and local self-attention mechanisms~\cite{wang2022spatial}. Following this, EfficientSCI~\cite{wang2023efficientsci} and EfficientSCI++~\cite{cao2024hybrid} employed dense connections and space-time factorization to improve reconstruction quality while reducing computational cost. 
A limitation of these E2E models is their lack of interpretability. Combining iterative optimization ideas with deep learning models, the deep unfolding networks can bring together the merits of both sides. For example, Wu et al. incorporated 3D-Unet as priors with dense feature map fusion to achieve excellent  performance~\cite{wu2021dense}. Yang et al. developed a novel ensemble learning-based prior for the unfolding network~\cite{yang2022ensemble}, which can outperform Dense3D-Unfolding~\cite{wu2021dense} with shorter inference time. All the above-mentioned reconstruction algorithms are designed for RS measurements. In this paper, we focus on USS measurements, where each sub-measurement is ultra-sparse sampled. To address this, we propose a sparse Transformer-based video SCI reconstruction method (dubbed BSTFormer), which is more suitable for USS measurements. As shown in Fig.~\ref{fig:psnr_time-1}, our proposed BSTFormer outperforms all previous deep learning-based reconstruction algorithms in terms of reconstruction quality with faster inference speed.

\subsection{Video Transformers}
\label{Sec:vit}
Transformers~\cite{vaswani2017attention} were initially developed for natural language processing. Recently, there are many explorations for introducing Transformers into video-related computer vision tasks~\cite{liu2022video, fu2022low,tu2019multi,wang2022bevt,wu2022survey,qing2023mar}. Video Swin Transformer~\cite{liu2022video} limited self-attention calculation to a local window, reducing computational complexity. However, they cannot well establish long-term correlations. TimeSformer~\cite{bertasius2021space} explored spatio-temporal relationships by separating spatial and temporal attentions. However, due to the use of global self-attention in the spatial domain, the computational complexity grows quadratically with image size. MaxViT~\cite{tu2022maxvit} introduced a multi-axis attention module composed of local block attention and global sparse attention, which allows efficient global-local spatial interactions with only linear complexity. Liang et al. proposed a recurrent video restoration Transformer (RVRT)~\cite{liang2022recurrent}, which processed neighboring frames in parallel using guided deformable attention within a globally recurrent framework. This design achieves a good balance between model size, effectiveness, and efficiency. XViT~\cite{bulat2021space} restricted temporal attention to a local temporal window and then approximated space-time attention with an efficient space-time mixing mechanism. ViViT~\cite{arnab2021vivit} served as a pure transformer-based model for video classification, where spatio-temporal tokens were extracted and efficient factorized versions of self-attention are applied to encode relationships between tokens. Zhang et al. integrated the completed optical flow into a flow-guided video Transformer for more accurate attention retrieval in the video inpainting tasks~\cite{zhang2022flow}. Uniformer~\cite{li2023uniformer} utilizes 3D CNN and spatial-temporal self-attention
respectively in shallow and deep layers, which allows tackling both redundancy and dependency for efficient representation learning. To reduce computational cost, Wang et al. proposed a token selection framework~\cite{wang2022efficient}, which can dynamically select more informative tokens in both spatial and temporal dimensions conditioned on input video samples. In this paper, unlike previous video Transformers, we propose a sparse video SCI reconstruction Transformer. Our model integrates local block attention, global sparse attention, and global temporal attention, which enables it to better exploit the sparsity in USS measurements for improved reconstruction quality.

\section{Preliminary}
\subsection{Mathematical Model of Video SCI}
\label{Sec:Model}
In this section, we present the mathematical model of video SCI. Let $\{\Xmat_m\}_{m=1}^{T}\in\mathbb{R}^{n_x\times{n_y}}$ denote the $T$-frame dynamic scene to be captured, where $n_x, n_y$ represent the spatial resolution of each frame and $T$ is the compression ratio (Cr). The modulation process, applied by DMD in our system as shown in Fig.~\ref{fig:video_R}, can be modeled as multiplying $\{\Xmat_m\}_{m=1}^{T}\in\mathbb{R}^{n_x\times{n_y}}$ by pre-defined masks $\{\Mmat_m\}_{m=1}^{T}\in\mathbb{R}^{n_x\times{n_y}}$, expressed as: 
\begin{equation}
    \Ymat_m=\Xmat_m\odot\Mmat_m, 
\end{equation}
where $\odot$ denotes the Hadamard (element-wise) multiplication, and $m\in\mathbb\{1,...,T\}$. Finally, the modulated dynamic scene is captured by a camera in a compressed manner. The forward model for the entire process can be expressed as: 
\begin{equation}
  \Ymat=\sum_{m = 1}^{T}{\Xmat_m\odot\Mmat_m+\Gmat},
  \label{eq:Y_mat}
\end{equation}
where $\Gmat\in\mathbb{R}^{n_x\times{n_y}}$ denotes measurement noise. 

Eq.~(\ref{eq:Y_mat}) can also be written in a vectorized form. First, we vectorize $\Ymat$, $\Gmat$, and $\Xmat$ as followings: $\yv=\operatorname{vec}(\Ymat)\in\mathbb{R}^{n_x{n_y}}$, $\gv=\operatorname{vec}(\Gmat)\in\mathbb{R}^{n_x{n_y}}$, and $\xv=\left[\xv_1^{\top},\ldots,\xv^{\top}_{T}\right]^{\top}\in\mathbb{R}^{n_x{n_y{T}}}$, where $\xv_m=\operatorname{vec}(\Xmat_m)$. The sensing matrix generated by the modulation masks $\{\Mmat_m\}_{m=1}^{T}$ is then defined as: 
  \begin{equation}
    \Phimat = \left[\Dmat_1,\ldots,\Dmat_{T}\right]\in\mathbb{R}^{n_x{n_y}\times{n_x{n_y{T}}}}, 
    \label{eq:H}
  \end{equation}
  where $\Dmat_m=\operatorname{Diag}(\operatorname{vec}(\Mmat_m)) \in \mathbb{R}^{n_{x} n_{y} \times n_{x} n_{y}}$ is a diagonal matrix with its diagonal elements filled by $\operatorname{vec}(\Mmat_m)$. 
  Finally, the vectorized form of Eq.~(\ref{eq:Y_mat}) can be written as:
  \begin{equation}
    \yv = \Phimat{\xv}+\gv. 
    \label{eq:y_vec}
  \end{equation}

In the decoding stage of the video SCI system, the desired video frames $\hat{\xv}$ can be reconstructed using a video SCI reconstruction algorithm after obtaining the compressed measurement $\yv$ and the modulation masks $\Phimat$.

\subsection{Differences Between the USS and RS Mask Sets}
\label{Sec:inp_sci_dif}
\begin{figure}[!htbp]
    \centering 
    \includegraphics[width=\linewidth]{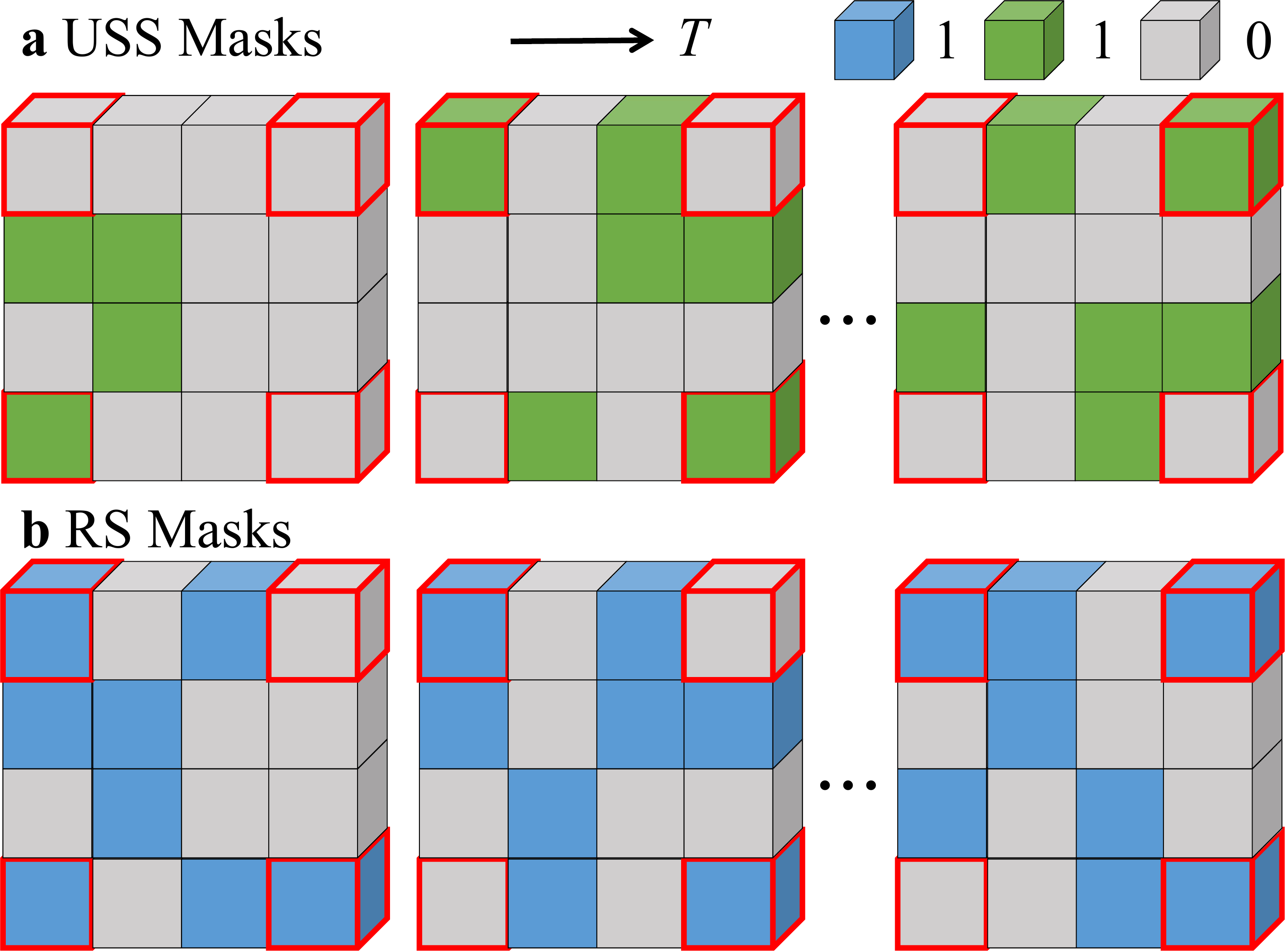}
    \caption{The difference between the USS and RS mask sets: a) As illustrated in four corners of the top USS mask cubes, at each spatial location, only one mask cube is set to 1 (green) and all others are set to 0 (gray). b) As shown in four corners of the bottom RS mask cubes, the color of each mask cube is randomly set to be 0 (gray) or 1 (blue) with no correlation.} 
  \label{fig:sci_inp_masks}
\end{figure}

In this section, let us give some details of the USS and RS mask sets. Please refer to Fig.~\ref{fig:sci_inp_masks} for a direct comparison between these two different mask designs.
\begin{itemize}
\item[a)] The RS mask set is a set of random binary masks with no correlation, which means that each mask element value is randomly set to be 0 or 1. 
\item[b)] The USS mask set is a binary mask set $\{\Mmat_m\}_{m=1}^{T} \in \mathbb\{0,1\}^{n_x \times n_y}$, which satisfies $\sum_{m=1}^{T} \Mmat_m = $\textbf{1}$^{n_x \times n_y}$, where $\textbf{1}$ is an all-one matrix. In this design, the masks are highly correlated. Specifically, for each spatial position in the mask, a single frame index $m \in \{1, \ldots, T\}$ is randomly chosen, where the mask value at this position is set to 1 for the $m$-th frame, and 0 for all other frames. 

Given this correlation, a USS measurement can be expressed as:
\begin{equation}
    \Ymat = \{\Ymat_1 \cup \Ymat_2 \cup \ldots \cup \Ymat_T\} \in \mathbb{R}^{n_x \times n_y},
\end{equation}
where $\Ymat_m$, $m \in \{1, \ldots, T\}$, represents the $m$-th inpainted (or masked) frame. Ideally, each inpainted frame can be decoupled using the relationship $\Ymat_m = \Ymat \odot \Mmat_m$. 
\end{itemize}

To illustrate the dynamic range of the USS masking strategy, we consider an 8-bit CCD where the intensity of each pixel ranges from 0 to 255 and assume that 10 frames are encoded and compressed into a snapshot measurement. For the RS mask set, at each pixel location, the energy from about 5 frames is integrated into one snapshot compressed measurement. This restricts the pixel intensity for each sub-frame to approximately [0, 51]. In contrast, for the USS mask set, only one frame is integrated into each snapshot compressed measurement at each pixel location, maintaining the full pixel intensity range of [0, 255]. Therefore, the USS masking strategy \textit{achieves a higher dynamic range} compared to the RS masking strategy.

\subsection{Self-Attention in Transformers}
Self-attention is a fundamental module in modern vision Transformer architectures and has consistently achieved SOTA performance on numerous computer vision tasks~\cite{liu2022video,wang2022bevt}. In this section, we briefly introduce the calculation process within the self-attention module. Given a sequence of $n$ entities $(\Xmat_1,\Xmat_2,...,\Xmat_n)$ expressed as $\Xmat\in\mathbb{R}^{n\times{d}}$, where $d$ is the embedding dimension of each entity. We first define three learnable weight matrices to transform Queries ($\Wmat^Q\in\mathbb{R}^{d\times{d_q}}$), Keys ($\Wmat^K\in\mathbb{R}^{d\times{d_k}}$), and Values ($\Wmat^V\in\mathbb{R}^{d\times{d_v}}$), where $d_q = d_k$. The input sequence is first projected onto these weight matrices $\Qmat=\Xmat{\Wmat^Q}$, $\Kmat=\Xmat{\Wmat^K}$, $\Vmat=\Xmat{\Wmat^V}$. The self-attention output $\Zmat\in\mathbb{R}^{n\times{d_v}}$ is computed as:
\begin{equation}
  \Zmat = softmax(\frac{\Qmat\Kmat^{T}}{\sqrt{d_q}}){\Vmat}.
  \label{eq:attention}
\end{equation}

For a given entity in the sequence, the self-attention mechanism calculates the dot product of its
query vector with all key vectors. This is followed by normalization using the softmax operator to compute the attention scores. Each entity is then updated as the weighted sum of all entities in the sequence, with the weights determined by the attention scores.
Based on the self-attention operation defined in Eq.~(\ref{eq:attention}), we propose local block attention, global sparse attention, and global temporal attention to enhance the performance of the USS-based video SCI reconstruction task.

\begin{figure*}[!ht]
    \centering 
    \includegraphics[width=\linewidth]{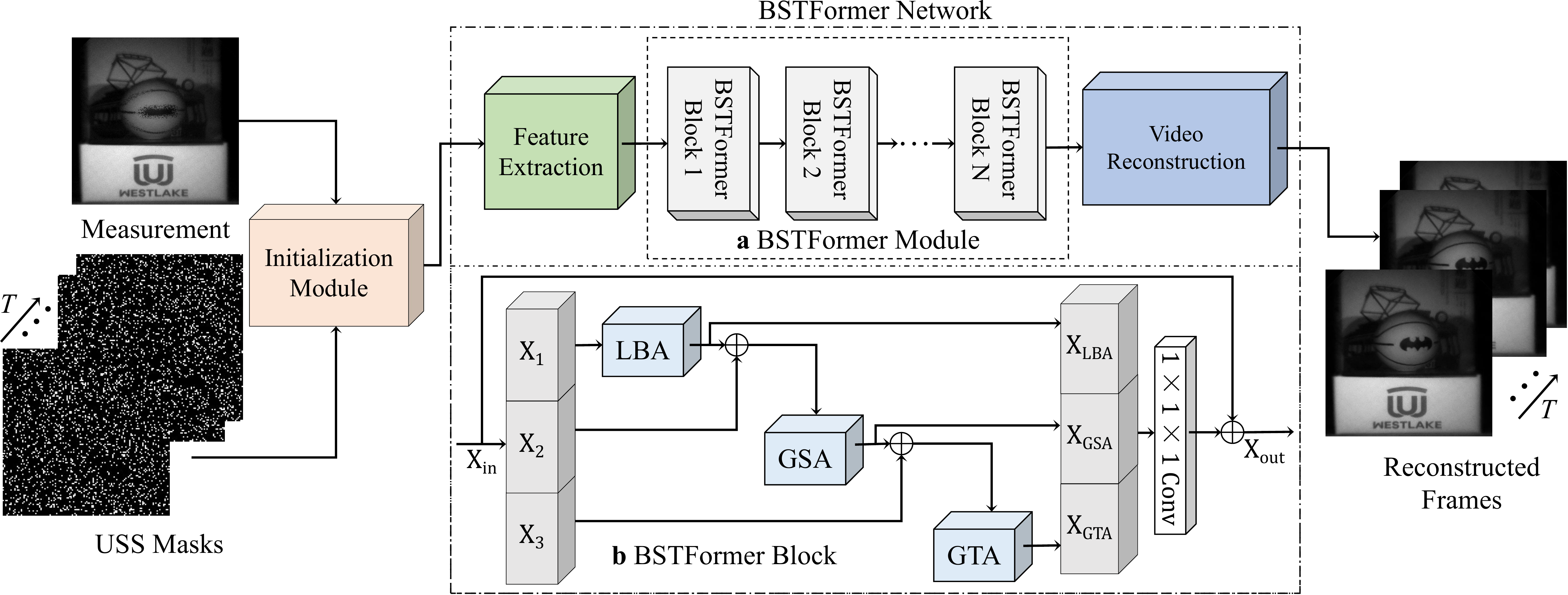}
    \caption{Illustration of our proposed method: Given a compressed measurement and the modulation masks, we first get the coarse estimated video frames by the initialization module. Then, the coarse estimated video frames are fed into the BSTFormer network (consists of a feature extraction module, a BSTFormer module, and a video reconstruction module) to obtain the final precise video frames. a) The BSTFormer module is composed of several BSTFormer blocks. b) The BSTFormer block contains a local block attention (LBA) branch, a global sparse attention (GSA) branch, and a global temporal attention (GTA) branch. These branches work together to enhance spatial-temporal feature representation and fully utilize the sparsity inherent in the USS measurement.}
    \label{fig:network}
\end{figure*}

\section{Proposed Reconstruction Network}
\label{Sec:recon}
In this section, we propose a novel reconstruction algorithm for the ultra-sparse sampled video SCI system, addressing the following two key considerations:
\begin{itemize}
\item[a)] As previously discussed, in the ideal case, the reconstruction problem can be viewed as an inpainting problem. Therefore, general image/video inpainting models can be used to reconstruct the ideal USS measurements. However, due to the mismatch between the DMD and CCD in the real-built video SCI system, the captured USS measurements cannot be directly decomposed. Keeping consistency in mind, we therefore propose a novel reconstruction algorithm for both ideal and real-world cases.
\item[b)] Existing video SCI reconstruction algorithms are primarily designed for random masking strategies. These methods fail to effectively leverage the inherent sparsity in the USS measurement, leading to suboptimal performance in the ultra-sparse sampling case.
\end{itemize}

As shown in Fig.~\ref{fig:network}, our proposed BSTFormer network consists of three main components: 
$i)$ feature extraction module, $ii)$ BSTFormer module, and $iii)$ video reconstruction module. Specifically, $i)$ The feature extraction module is composed of 
three 3D convolutional layers with kernel sizes of $3\times{7}\times{7}$, $3\times{3}\times{3}$, and $3\times{3}\times{3}$. 
Each convolutional layer is followed by a LeakyReLU activation function~\cite{maas2013rectifier}. The final 3D convolution uses a spatial stride of $1\times{2}\times{2}$, reduces the spatial resolution of the output feature map to half of the input resolution. 
The feature extraction module can effectively map the input image space to a high-dimensional feature space. In the BSTFormer configuration, the module has 1 input channel and 192 output channels. $ii)$ The BSTFormer module contains 4 BSTFormer blocks, which can efficiently explore spatial-temporal correlation and sparsity within the USS measurement. In our experiments, both the input and output channels of this module are set to 192. $iii)$ The video reconstruction module consists of a transposed 3D convolutional layer with a kernel size of $1 \times 3 \times 3$ to restore the spatial resolution to the input size. It also includes two additional 3D convolutional layers with kernel sizes of $1 \times 1 \times 1$ and $3 \times 3 \times 3$. These layers reconstruct video frames from the features output by the BSTFormer module. The input and output channels of this module are set to 192 and 1, respectively.

\subsection{Initialization Module}
As shown in Fig.~\ref{fig:network}, the initialization module of the proposed BSTFormer network is inspired by~\cite{cheng2020birnat,cheng2021memory}. In this module, the measurement ($\Ymat$) and masks ($\Mmat$) are processed as follows:
\begin{equation}
  \overline{\Ymat}=\Ymat\oslash \sum_{m = 1}^{T}{\Mmat_m}, \;
  ~~X_{e,m} = \overline{\Ymat} \odot {\Mmat_m} + \overline{\Ymat}, 
  \label{eq:Y_line}
\end{equation}
where $\oslash$ represents Hadamard (element-wise) division, 
$\overline{\Ymat}\in\mathbb{R}^{n_x\times{n_y}}$ is the normalized measurement, which preserves a certain degree of the background and motion trajectory information, 
and $\Xmat_{e,m}\in\mathbb{R}^{n_x\times{n_y}}$ represents the coarse estimation of the $m$-th desired video frame. 
We then take ${\{\Xmat_{e,m}}\}_{m=1}^{T}$ as the input of the proposed network to get the final reconstruction result. 
Note that, in the ideal case, the proposed USS masks can be summed up to an all-one matrix, so that dividing by $\sum_{m = 1}^{T}{\Mmat_m}$ has no effect. However, due to the mismatch between the DMD and CCD in real-built video SCI systems, the USS measurements cannot be ideally decomposed. In such cases, some pixels are summed up to be less than one, while others exceed one. Thus, in order to maintain consistency, we apply Eq.~(\ref{eq:Y_line}) to both the ideal and real-world scenarios.

The reasoning behind the input processing in Eq.~(\ref{eq:Y_line}) can be summarized as follows: From Eq.~(\ref{eq:Y_mat}), we know that the measurement $\Ymat$ is a weighted summation of the dynamic video frames. Consequently, the real-captured measurement $\Ymat$ is typically not energy-normalized. For example, some pixels in the real-captured measurement $\Ymat$ may accumulate less than one-pixel energy, while others may accumulate more. This non-uniform energy distribution makes it unsuitable to directly feed $\Ymat$ into the reconstruction network. Therefore, we apply a measurement energy normalization method, as shown in the left half of Eq.~(\ref{eq:Y_line}), to address this issue.

\begin{figure*}[!htbp]
    \centering 
    \includegraphics[width=.75\linewidth]{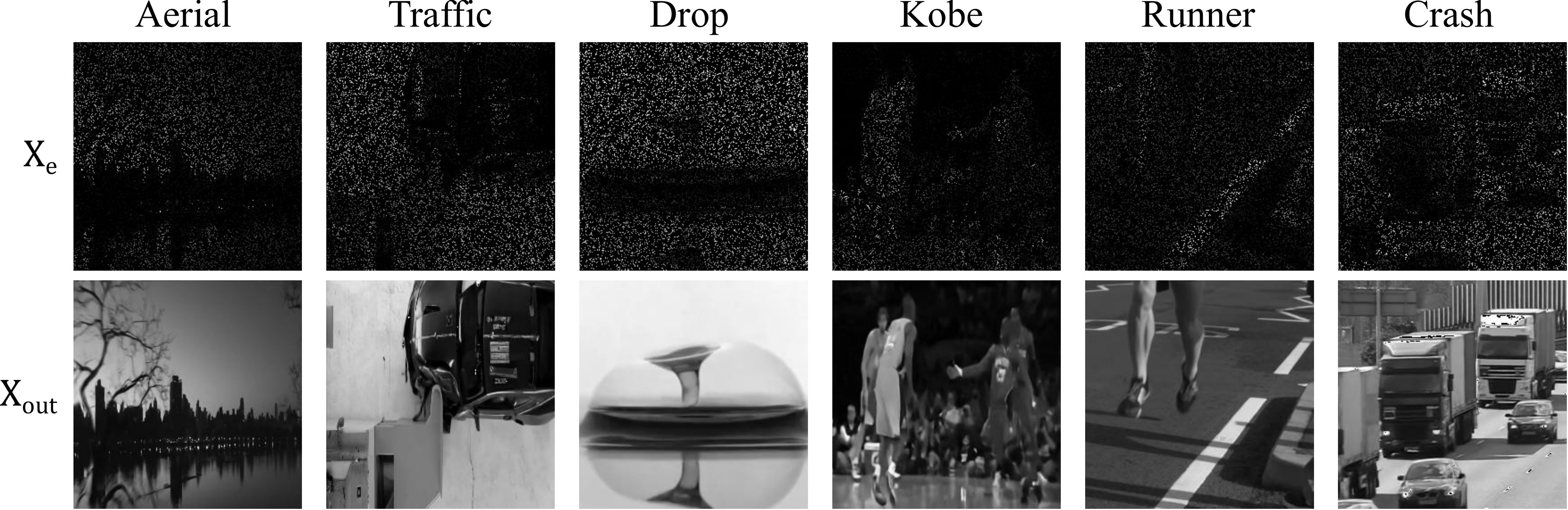}
    \caption{Initialized video frame $\Xmat_e$ and final reconstructed video frame $\Xmat_{out}$ of our BSTFormer network when testing on the simulated USS data.} 
  \label{fig:xe_main}
\end{figure*}

Finally, we plot the coarse estimation $\Xmat_e$ and the final reconstructed video $\Xmat_{out}$ in Fig.~\ref{fig:xe_main}. It can be clearly observed that the reconstruction quality significantly improves after applying our BSTFormer. This demonstrates the superior performance of our
BSTFormer.   

~~\\
\subsection{BSTFormer Block}
As shown in Fig.~\ref{fig:network}, the BSTFormer block is designed to efficiently capture spatial-temporal correlations while reducing computational cost. The input features are first split along the channel dimension. Each split is then processed through one of three attention branches: Local Block Attention (LBA), Global Sparse Attention (GSA), and Global Temporal Attention (GTA).
To enhance feature fusion and propagation, skip connections are incorporated between adjacent attention blocks. These connections help preserve critical information and stabilize feature learning. The outputs of the three attention branches are concatenated and passed through a $1 \times 1 \times 1$ 3D convolutional layer, which generates the final output of the BSTFormer block.
Below, we provide detailed descriptions of the three attention branches.

\begin{figure}[t]
    \centering
    \includegraphics[width=\linewidth] {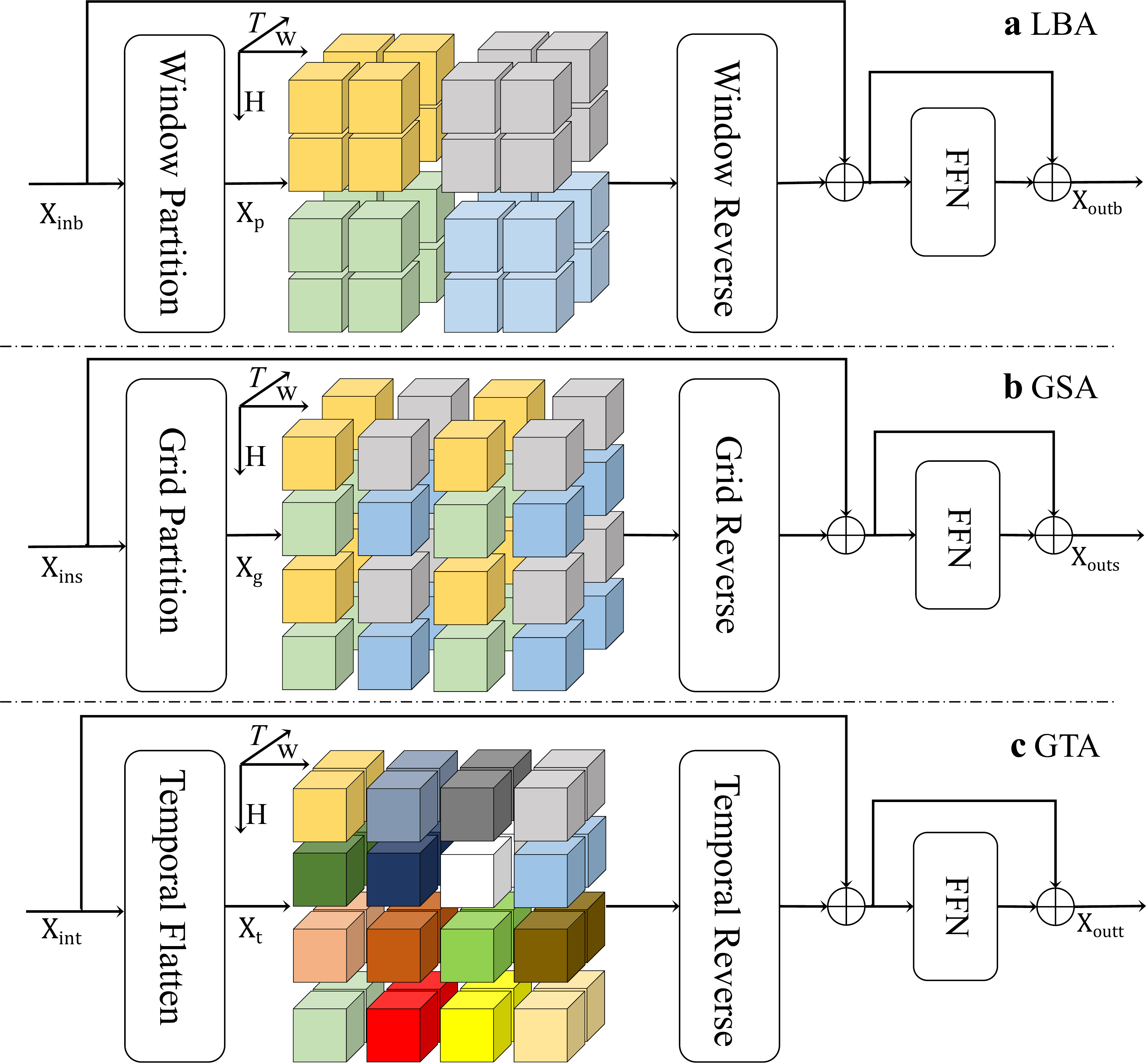}
    \caption{The structure of LBA, GSA, and GTA. a) In the LBA branch, the input feature is divided into several non-overlapping windows to focus on local spatial features. b) In the GSA branch, the input feature is partitioned into uniform grids. c) In the GTA branch, self-attention calculations are performed at each spatial location across the entire temporal dimension. For simplicity, window size and grid size are both $2\times2$, temporal length is also 2.}
    \label{fig:attention}
\end{figure}

In the spatial domain, we first utilize LBA to extract local features, as shown in Fig.~\ref{fig:attention}. Consider an input feature map $\Xmat_{inb}\in\mathbb{R}^{T\times W\times H\times C}$, where $T,W,H,C$ represent Cr, width of each frame, height of each frame, and channel number, respectively. The feature map $\Xmat_{inb}$ is divided into several non-overlapping windows of size $S \times S$ to obtain the window-partitioned input feature $\Xmat_{p} \in \mathbb{R}^{L \times J \times C}$, where $J = S \times S$ is the number of tokens in each local window, and $L = T \times \frac{W \times H}{S \times S}$ represents the number of local windows. In our network, $S$ is set to 7. The self-attention calculation in Eq.~(\ref{eq:attention}) is then applied to each non-overlapping local window, producing the output feature $\Xmat_{outb}$.

Next, considering the sparsity within the USS measurement, a large proportion of information is lost in each sub-measurement. Thus, it is beneficial to explore global correlations in the spatial domain. To this end, we adopt GSA to establish global spatial correlations following the LBA branch. As shown in Fig.~\ref{fig:attention}, the input feature map $\Xmat_{ins} \in \mathbb{R}^{T \times W \times H \times C}$ is divided into several uniform grids of size $\frac{W}{G} \times \frac{H}{G}$, resulting in the grid-partitioned input feature $\Xmat_{g} \in \mathbb{R}^{L \times J \times C}$, where $J = T \times \frac{W \times H}{G \times G}$ is the number of tokens in each grid, and $L = G \times G$ represents the number of uniform grids. In our network, $G$ is set to 7. The self-attention operator is then applied to each uniform grid, generating the output feature $\Xmat_{outs}$. 

In the temporal domain, GTA is used to establish long-range temporal relationships. Specifically, as shown in Fig.~\ref{fig:attention}, the input feature map $\Xmat_{int} \in \mathbb{R}^{T \times W \times H \times C}$ is reshaped into $\Xmat_{t} \in \mathbb{R}^{WH \times T \times C}$. The self-attention calculation is then performed along the temporal dimension at each spatial location, resulting in the output feature $\Xmat_{outt}$.  

Finally, as shown in Fig.~\ref{fig:attention}, at the end of each attention branch, a Feed Forward Network (FFN), consisting of two 3D convolutional layers, is used to further integrate spatial-temporal information. Specifically, given $\Xmat_f \in \mathbb{R}^{T \times H \times W \times C}$, the FFN is defined as:
\begin{equation}
  \hat{\Xmat}_f = \Xmat_f + \Wmat_1(\phi(\Wmat_2(\Xmat_f))),
\end{equation}
where $\Wmat_1$ and $\Wmat_2$ represent $1 \times 1 \times 1$ and $3 \times 3 \times 3$ convolutional operations, respectively. $\phi$ denotes the LeakyReLU activation function, and $\hat{\Xmat}_f \in \mathbb{R}^{T \times H \times W \times C}$ is the output of the FFN. 

\noindent{\bf Comparison with prior spatial-temporal Transformers for video SCI reconstruction:} Spatial-temporal Transformers are widely used in video-related tasks~\cite{geng2022rstt,zhou2022transvod,gao2023mist} due to their superior performance. So far, several spatial-temporal Transformers have been proposed for video SCI reconstruction, including STFormer~\cite{wang2022spatial}, EfficientSCI~\cite{wang2023efficientsci}, and EfficientSCI++~\cite{cao2024hybrid}. Specifically, $i$) In the STFormer netowrk~\cite{wang2022spatial}, window-based self-attention mechanism is used to explore spatial and temporal correlations, respectively. $ii$) In the EfficientSCI~\cite{wang2023efficientsci} and EfficientSCI++~\cite{cao2024hybrid} methods, a hybrid CNN-Tranformer block
is built, which can efficiently establish space-time correlations by using convolution in the spatial domain and
Transformer in the temporal domain. $iii$) Unlike previous spatial-temporal Transformers for video SCI reconstruction, our BSTFormer incorporates local block attention, global sparse attention, and global temporal attention. Among these, global sparse attention is specially design to explore the sparsity within USS data, as verified through experiments.

\subsection{Computational Complexity}
In this section, we analyze the computational complexity of our proposed BSTFomer network (${\rm BSTF}$). 
First, the computational complexity of the Multi-head Self-Attention mechanism (MSA) in Transformer can be expressed as:
\begin{align}
\varOmega({\rm MSA})&=4SC^2+2S^{2}C, 
\end{align}
where $\varOmega(\cdot)$ represents the computational complexity, $S$ represents the sequence length, and $C$ is the channel number. 
In the LBA module, the sequence length of a single grid  is $G^2$, the channel number is $\frac{C}{3}$, and the number of grids is $\frac{HWT}{G^2}$. Thus, the computational complexity of the LBA module can be expressed as:
\begin{align}
\varOmega({\rm LBA})&=(\frac{4G^2C^2}{9}+\frac{2G^4C}{3})\times{\frac{HWT}{G^2}} \notag\\
&=\frac{4}{9}HWTC^2+\frac{2}{3}G^2HWTC.
\end{align}
Similarly, the computational complexity of the GSA module and the GTA module can be expressed as:
\begin{align}
\varOmega({\rm GSA})&=\frac{4}{9}HWTC^2+\frac{2}{3}G^2HWTC, \\
  \varOmega({\rm GTA})&=\frac{4}{9}HWTC^2+\frac{2}{3}HWT^{2}C.
\end{align} 
Finally, the total computational complexity of the proposed BSTFormer network can be calculated as:
\begin{align}
\varOmega({\rm BSTF})&=\varOmega({\rm LBA}) +\varOmega({\rm GSA})+\varOmega({\rm GTA})  \notag\\
  &=\frac{4}{3}HWTC^2+\frac{4}{3}G^2HWTC \notag\\
  &\quad+\frac{2}{3}HWT^{2}C, 
\end{align} 
where $G$ and $C$ are pre-defined constants that do not change with the input size. $T$ is the compression ratio, which is smaller than $HW$. Therefore, the computational complexity of the proposed BSTFormer network grows linearly with $HW$, requiring significantly less computation than ${\rm G\mbox{-}MSA}$, whose complexity is quadratic with respect to $HW$:
\begin{align}
\varOmega({\rm G\mbox{-}MSA}) &= 4HWTC^2 + 2(HWT)^2C.
\end{align}

\section{Experimental Results}
\label{Sec:Results}

\begin{figure}[!ht]
    \centering   \includegraphics[width=\linewidth] {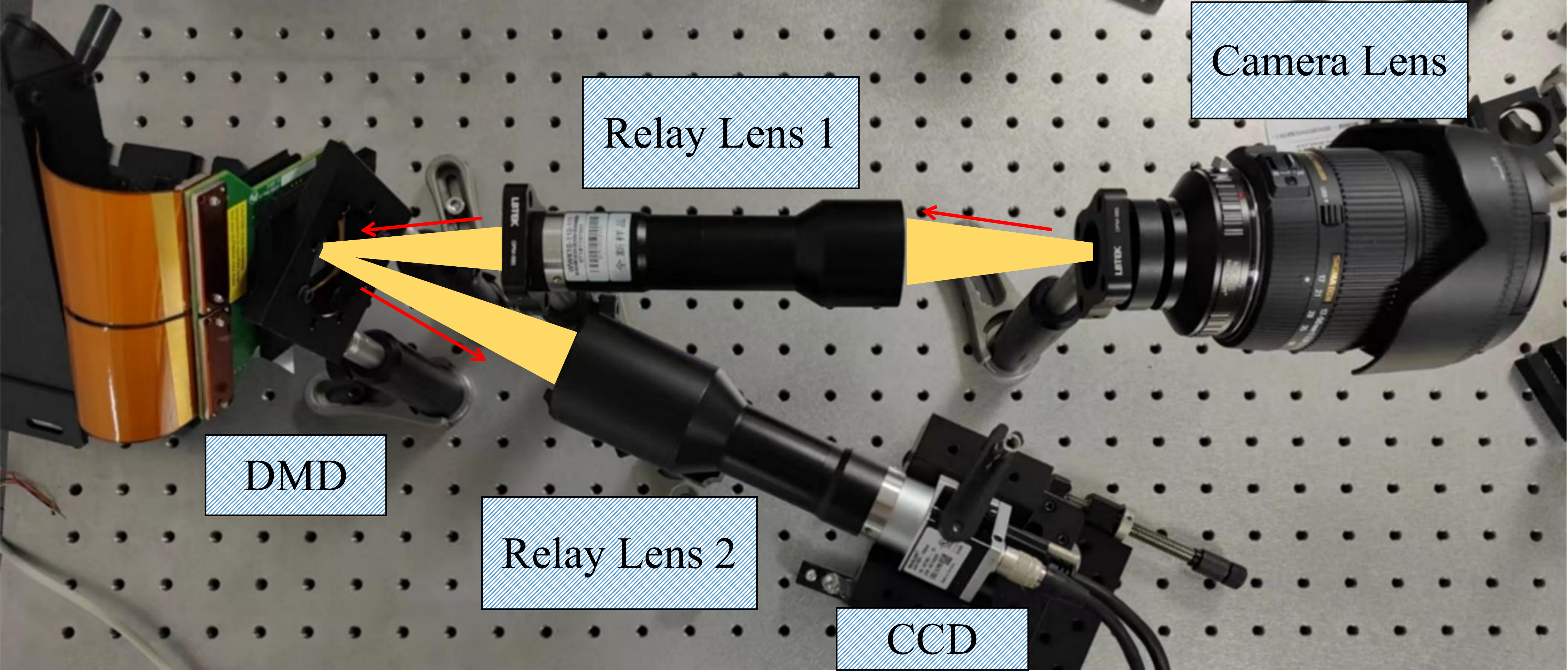}
    \caption{Real-built video SCI system. The encoding process is as follows: First, the reflected light from the target scene is imaged onto the surface of DMD via a camera lens. Then, the dynamic scene is modulated by the pre-stored patterns in the DMD. Finally, the encoded measurement is projected onto a CCD through a relay lens during each single exposure time.}
    \label{fig:video_R}
\end{figure}

\noindent{\bf Hardware Implementation:} The optical setup of the video SCI system is shown in Fig.~\ref{fig:video_R}. The encoding process is as follows: First, the reflected light from the target scene is imaged onto the surface of DMD (TI, $2560 \times 1600$ pixels, $7.6 \mu m$ pixel pitch) via a camera lens (Sigma, 17-50/2.8, EX DC OS HSM) and a relay lens (Coolens, WWK10-110-111). It is worth mention that DMD is a modulation device which can rapidly switch between different masks at high frequencies. Next, the projected dynamic scene is modulated by the USS or RS masks, which are pre-stored in the DMD. Finally, the modulated scene is projected onto a CCD (Basler acA1920, $1920 \times 1200$ pixels, $4.8 \mu m$ pixel pitch) with another relay lens (Coolens, 
WWK066-110-110). In our experiments, the bit depth of the CCD is 8, meaning that the pixel intensity ranges from 0 to 255. In this way, the CCD can capture a series of compressed measurements in a snapshot manner.

\noindent{\bf Datasets:} Three types of datasets are used in this paper: training dataset, simulated testing dataset, and real-world testing dataset. $i$) Following BIRNAT~\cite{cheng2020birnat}, we first choose the \texttt{DAVIS2017} dataset~\cite{pont20172017} with spatial resolution $480\times{894}$ as the training dataset. $ii$) We then test our BSTFormer network on six simulated testing datasets, including \texttt{Kobe, Traffic, Runner, Drop, Crash,} and \texttt{Aerial} (with spatial resolution $256\times{256}$ and Cr 8). For the simulated testing datasets, Peak Signal-to-Noise-Ratio (PSNR) and Structural SIMilarity (SSIM)~\cite{wang2004image} are used as the evaluation metrics of the reconstruction quality. $iii$) Finally, we test the proposed BSTFormer network on four real-world testing datasets, including \texttt{Rotating Ball}, \texttt{Rotating Windmill}, \texttt{Falling Domino}, and \texttt{Waving Hand} (with spatial resolution $800\times{800}$ and Crs \{10, 20, 30, 40, 50\}) captured by our real-built video SCI system. 

\noindent{\bf Training Details:} We train the proposed network with the DAVIS2017 dataset on 4 NVIDIA RTX 3090 GPUs. First, regular data augmentation operations such as random cropping, random scaling, and random horizontal flipping are performed on the training data. Then, we can get the desired SCI measurements according to the forward model of the video SCI system described in Eq.~(\ref{eq:Y_mat}). Finally, we set Cr as 8 and send the generated measurements along with a set of fixed masks into the network for training with Adam optimizer~\cite{kingma2015adam}. Specifically, we first train for 200 epochs with initial learning rate 0.0001 on the training dataset with a spatial size of $128\times128$ and then fine-tune the network for 30 epochs with initial learning rate 0.00001 with a spatial size of $256 \times 256$.

Following EfficientSCI~\cite{wang2023efficientsci}, we choose the mean square error (MSE) as loss function. Our proposed method takes the measurement (\Ymat) and the corresponding masks ($\{\Mmat_m\}_{m=1}^{T}$) as inputs, and then generates the reconstructed video frames ($\{\hat{\Xmat}_m\}_{m=1}^{T}\in\mathbb{R}^{n_x\times{n_y}}$). Therefore, the loss function can be defined as:
\begin{equation}
\mathcal{L}_{MSE}= \textstyle \frac{1}{Tn_xn_y}\sum_{m = 1}^{T}\Vert \hat{\Xmat}_m - \Xmat_m \Vert_{2}^{2},
\end{equation}
where $\Xmat_m$ is the ground truth.

\begin{figure*}[!ht]
    \centering
    \includegraphics[width=.75\linewidth] {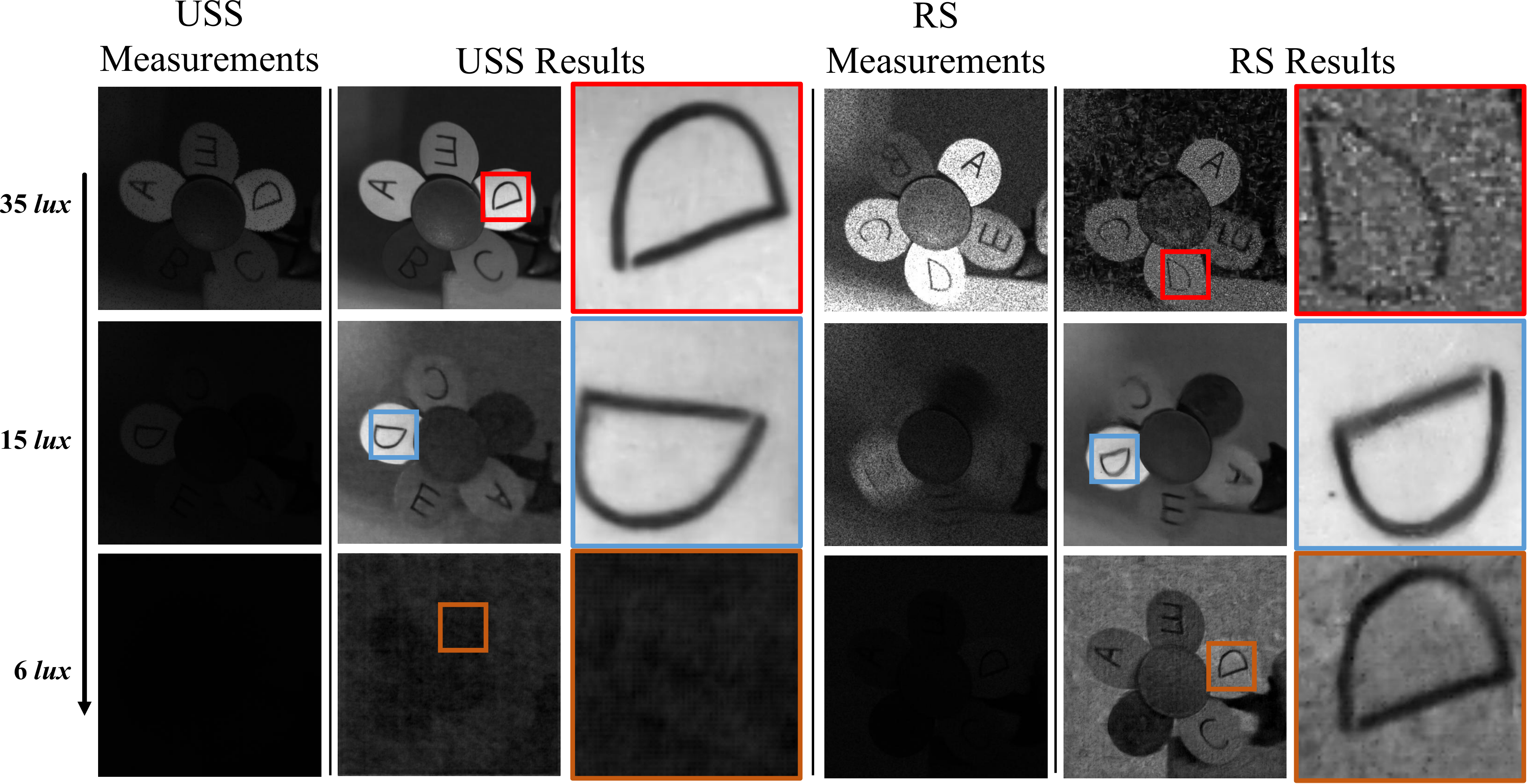}
    \caption{Reconstructed frames of the real-world RS and USS measurements under the same lighting conditions from 6$lux$ to 35$lux$ in our experimental setting. Cr is 10 in this case. See \textcolor{blue}{Supplementary Movie 1} for the complete video.}
    \label{fig:RS_USS_hdr_cr10}
\end{figure*}

\begin{figure}[!ht]
    \centering
    \includegraphics[width=\linewidth] {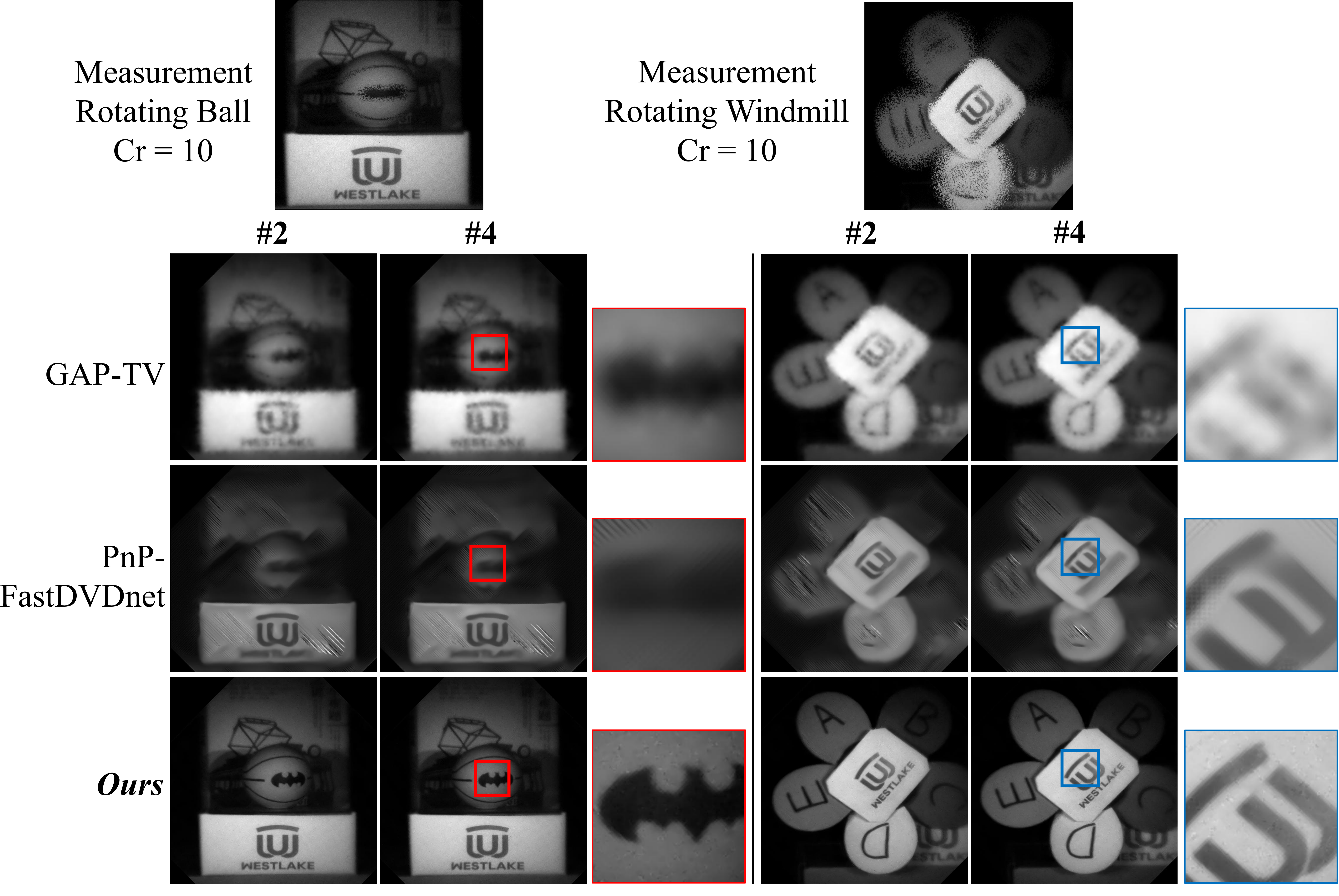}
    \caption{Reconstructed frames of the real-world USS measurements with different methods (including GAP-TV~\cite{Yuan16ICIP_GAP}, PnP-FastDVDnet~\cite{yuan2021plug}, and our BSTFormer) when Cr is 10. See \textcolor{blue}{Supplementary Movie 2} for the complete video.}
    \label{fig:RS_USS_cr10}
\end{figure}

\subsection{Results on Real-world Data}
\begin{figure}[!ht]
    \centering
    \includegraphics[width=\linewidth] {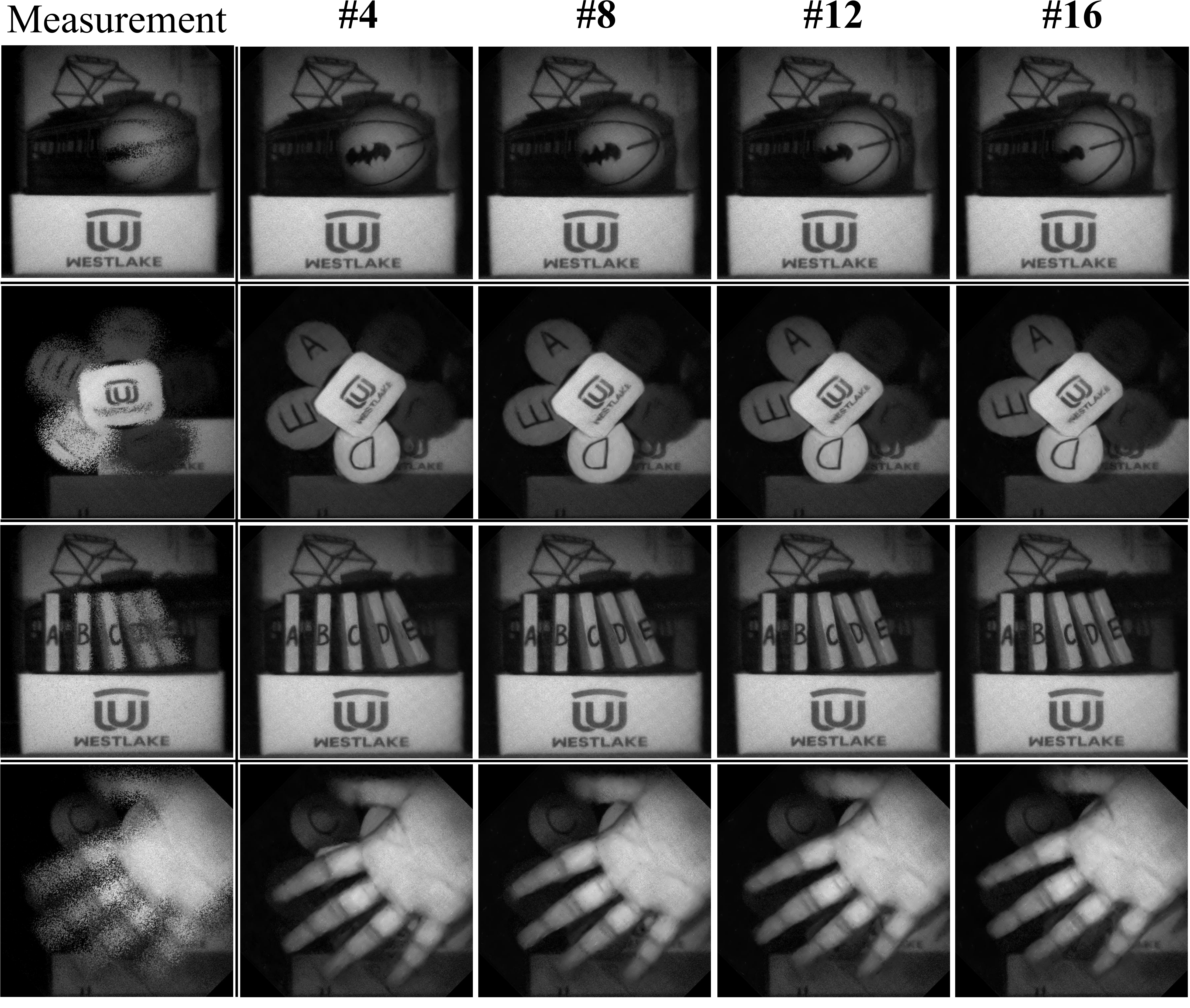}
    \caption{Reconstructed frames of the real-world USS measurements using our proposed BSTFormer when Cr is 20. See \textcolor{blue}{Supplementary Movie 3} for the complete video.}
    \label{fig:RS_USS_cr20}
\end{figure}

\begin{figure}[!ht]
    \centering
    \includegraphics[width=\linewidth] {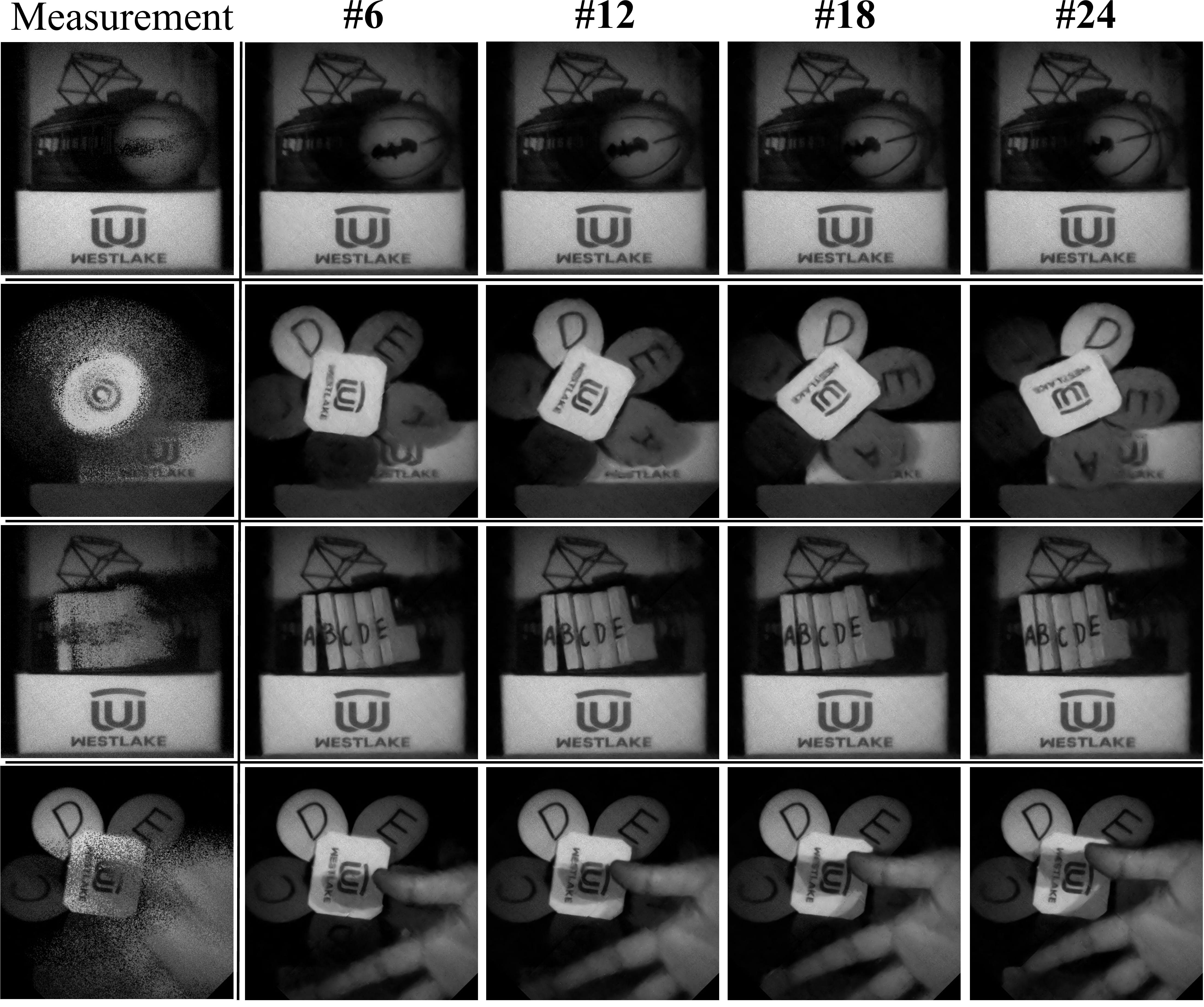}
    \caption{Reconstructed frames of the real-world USS measurements using our proposed BSTFormer when Cr is 30. See \textcolor{blue}{Supplementary Movie 4} for the complete video.}
    \label{fig:RS_USS_cr30}
\end{figure}

\begin{figure}[!ht]
    \centering
    \includegraphics[width=\linewidth] {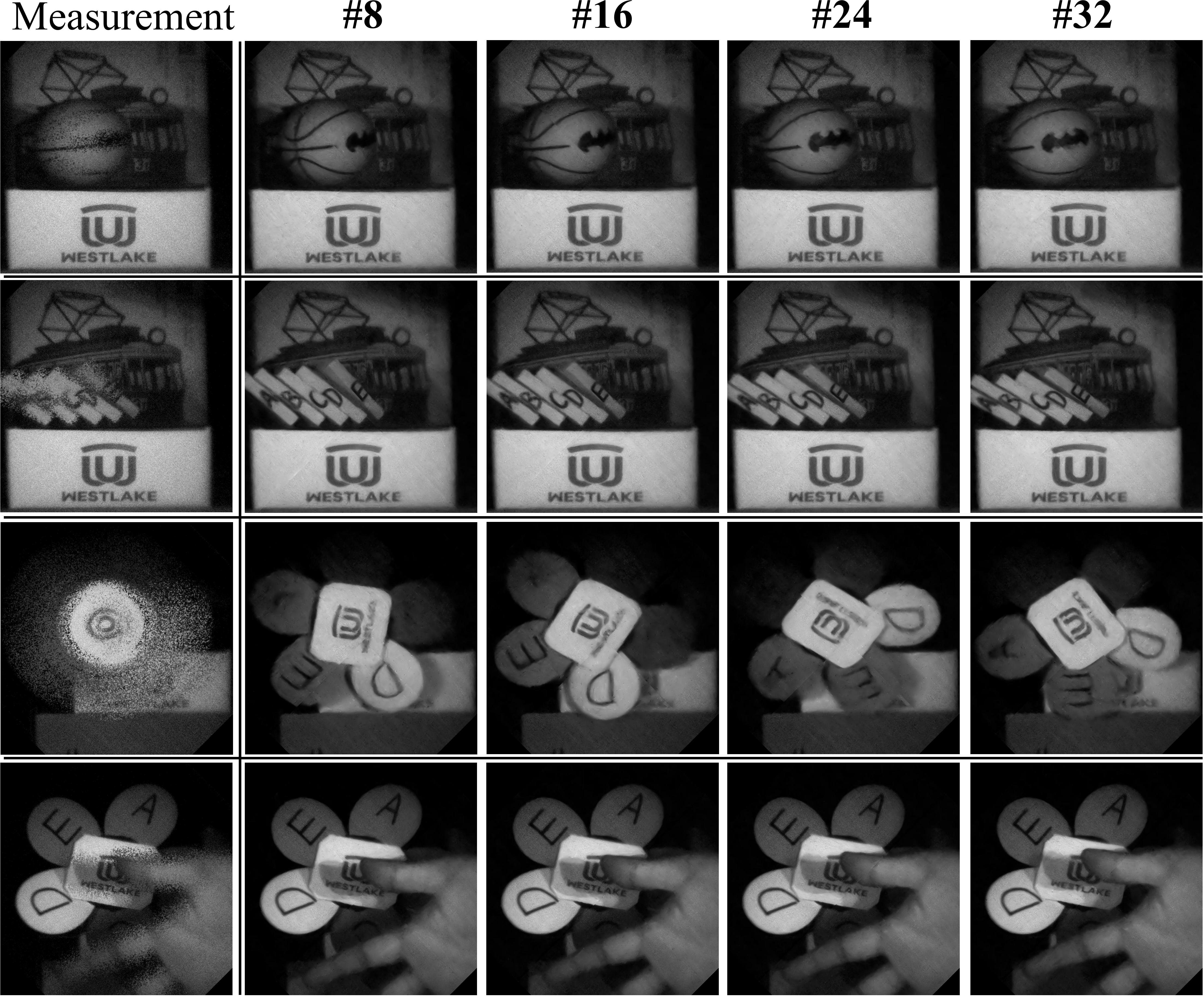}
    \caption{Reconstructed frames of the real-world USS measurements using our proposed BSTFormer when Cr is 40. See \textcolor{blue}{Supplementary Movie 5} for the complete video.}
    \label{fig:RS_USS_cr40}
\end{figure}

\begin{figure}[!ht]
    \centering
    \includegraphics[width=\linewidth] {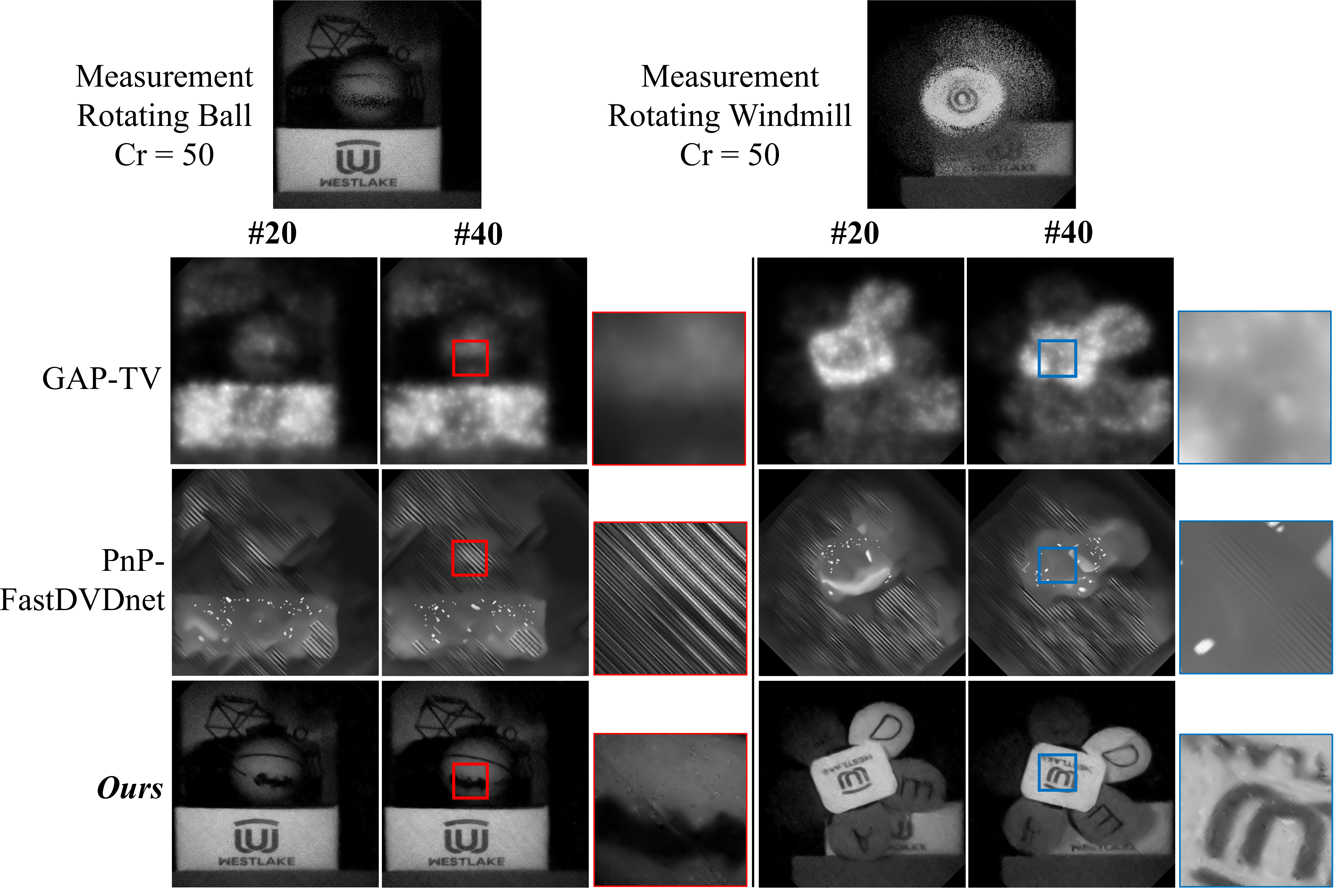}
    \caption{Reconstructed frames of the real-world USS measurements with different methods (including GAP-TV~\cite{Yuan16ICIP_GAP}, PnP-FastDVDnet~\cite{yuan2021plug}, and our proposed BSTFormer algorithm) when Cr is 50. See \textcolor{blue}{Supplementary Movie 6} for the complete video.}
    \label{fig:RS_USS_cr50}
\end{figure}

In this section, we compare the reconstruction quality of the real-world RS and USS measurement under different lighting conditions. For a fair comparison, we obtain the reconstructed frames of the real-world RS and USS measurements using EfficientSCI (SOTA method for RS measurement) and our proposed BSTFormer (SOTA method for USS measurement). We also evaluate real-world USS measurements with different Crs ranging from 20 to 50. With the 2D camera operates at 50$fps$, the equivalent frame rate of our real-built video SCI system is {500}$fps$ and {2500}$fps$ when Cr is 10 and 50, respectively. Due to noise uncertainties in the real-built video SCI system and the imperfect match between the DMD and CCD, reconstructing real-captured measurements—particularly those with ultra-sparse sampling—poses significant challenges. It is worth noting that the spatial size of the real-world testing data is $800\times800$, and Crs = \{10, 20, 30, 40, 50\}. To the best of our knowledge, this is the first real-world USS dataset with such a large spatial resolution and diverse Cr settings. To better compare reconstruction quality, we zoom in on selected local areas of the recovered frames. Moreover, since the same light intensity is used to capture both RS and USS measurements, the USS measurements appear very dark due to low light efficiency, as shown in Fig.~\ref{fig:RS_USS_hdr_cr10}. To better visualize details, we normalize the USS measurements by scaling all pixels to their own maximum values. For simplicity, only the normalized USS measurements are shown from Fig.~\ref{fig:RS_USS_cr10} to Fig.~\ref{fig:RS_USS_cr50}.

We can summarize the observations as follows:  
\begin{itemize}
\item[1)] From Fig.~\ref{fig:RS_USS_hdr_cr10}, we can see that when the illumination intensity is strong (35$lux$ in our experimental setting), the real-captured RS measurement becomes over-exposed, leading to the failure of reconstruction. In contrast, the real-captured USS measurement can still be normally exposed, allowing for accurate reconstruction. This demonstrates that the dynamic range of the real-world USS measurement is much higher than that of the RS one, highlighting a critical advantage of the proposed USS masking strategy.
\item[2)] We can learn from Fig.~\ref{fig:RS_USS_hdr_cr10} that under normal lighting conditions (15$lux$ in our experiments), both the RS and USS measurements are properly exposed. In such cases, our BSTFormer produces USS results with competitive quality compared to the RS results, demonstrating the superior performance of our BSTFormer.
\item[3)] We can observe from Fig.~\ref{fig:RS_USS_hdr_cr10} that under low lighting conditions (6$lux$ in our experiments), we can obtain correct RS results, despite the presence of noticeable noise. In contrast, due to the limited light efficiency of the proposed USS masking strategy, the reconstruction process fails. Therefore, light efficiency remains a challenge in the proposed USS masking strategy. 
\item[4)] From Fig.~\ref{fig:RS_USS_cr10}, we observe that our BSTFormer can recover clearer scenes and sharper edges than GAP-TV and PnP-FastDVDnet. This is particularly noticeable in the Batman logo on the \texttt{Rotating Ball} and the letters on the \texttt{Rotating Windmill} when Cr is 10.
\item[5)] From Fig.~\ref{fig:RS_USS_cr20} to Fig.~\ref{fig:RS_USS_cr40}, the proposed BSTFormer consistently produces high-quality reconstructed video frames as Cr increases from 20 to 40.
\item[6)] Fig.~\ref{fig:RS_USS_cr50} shows that even when Cr reaches a high value of 50, the proposed BSTFormer can also give us a better reconstruction quality than previous methods. Additionally, the reconstruction time for a real-captured measurement (Cr = 50) is approximately 42 minutes for GAP-TV and 22 minutes for PnP-FastDVDnet, whereas the BSTFormer requires only about 4.5 seconds, making it significantly faster than GAP-TV and PnP-FastDVDnet.
\end{itemize}

\begin{table*}[!htbp]
  \renewcommand{\arraystretch}{1.0}
  \caption{The average PSNR in dB (left entry), SSIM (right entry) and inference time per measurement of different video SCI reconstruction algorithms when testing on the simulated USS data. 
  The best results are shown in bold and the second-best results are underlined.}
  \centering
  \resizebox{\textwidth}{!}
  {
  \centering
  \begin{tabular}{c|c|c|c|c|c|c|c|c}
  \hline
  Dataset 
  & Kobe 
  & Traffic 
  & Runner 
  & Drop 
  & Crash 
  & Aerial 
  & Average 
  & Inference time (s) 
  \\
  \hline
  \hline
  GAP-TV~\cite{Yuan16ICIP_GAP}
  & 23.18, 0.735
  & 18.36, 0.522
  & 24.07, 0.760
  & 26.94, 0.828 
  & 21.91, 0.666 
  & 22.02, 0.685 
  & 22.75, 0.699
  & 4.2 (CPU) \\
  \hline 
  PnP-FFDNet~\cite{yuan2020plug} 
  & 23.38, 0.789  
  & 19.88, 0.710
  & 25.80, 0.872
  & 32.67, 0.968
  & 22.43, 0.780
  & 22.04, 0.753
  & 24.28, 0.812
  & 3.0 (GPU)
  \\
  \hline
  PnP-FastDVDnet~\cite{yuan2021plug} 
  & 26.21, 0.863
  & 24.78, 0.865 
  & 32.59, 0.942 
  & 38.71, 0.984 
  & 25.46, 0.883 
  & 26.57, 0.869
  & 29.05, 0.901
  & 6.0 (GPU)
  \\
  \hline
  BIRNAT~\cite{cheng2020birnat} 
  & 28.58, 0.900
  & 26.94, 0.922 
  & 35.76, 0.970
  & 40.26, 0.989
  & 26.81, 0.903
  & 27.29, 0.897
  & 30.94, 0.930
  & 0.16 (GPU)
  \\
  \hline
  RevSCI~\cite{cheng2021memory}
  & 27.45, 0.854
  & 26.99, 0.921
  & 36.00, 0.970
  & 40.47, 0.989
  & 26.69, 0.896
  & 27.64, 0.903
  & 30.87, 0.922
  & 0.19 (GPU)
  \\
  \hline
  STFormer-B~\cite{wang2022spatial}
  & 30.55, 0.946
  & 29.42, 0.952
  & \underline{38.76}, \underline{0.983}
  & 42.04, 0.992
  & 28.72, 0.950
  & \underline{29.79}, \underline{0.938}
  & 33.21, 0.960
  & 0.49 (GPU)
  \\
  \hline
  EfficientSCI-B~\cite{wang2023efficientsci}
  & \underline{30.83}, \underline{0.949}
  & \underline{29.60}, \underline{0.955}
  & 38.50, 0.983
  & \underline{42.34}, \underline{0.992}
  & \underline{29.05}, \underline{0.955}
  & 29.53, 0.936
  & \underline{33.31}, \underline{0.962}
  & 0.31 (GPU)
  \\
  \hline
  \rowcolor{lightgray}
  Our BSTFormer
  & {\bf 32.61}, {\bf 0.962}
  & \bf{29.82}, \bf{0.955}
  & {\bf 39.77}, {\bf 0.984}
  & {\bf 43.94}, {\bf 0.993}
  & {\bf 29.20}, {\bf 0.955}
  & {\bf 30.04}, {\bf 0.939}
  & {\bf 34.23}, {\bf 0.965}
  & 0.22 (GPU)
  \\
  \hline 
  \end{tabular}
  }
  \label{Tab:sim6}
\end{table*}

\begin{figure*}[!htbp]
    \centering
    \includegraphics[width=.7\linewidth] {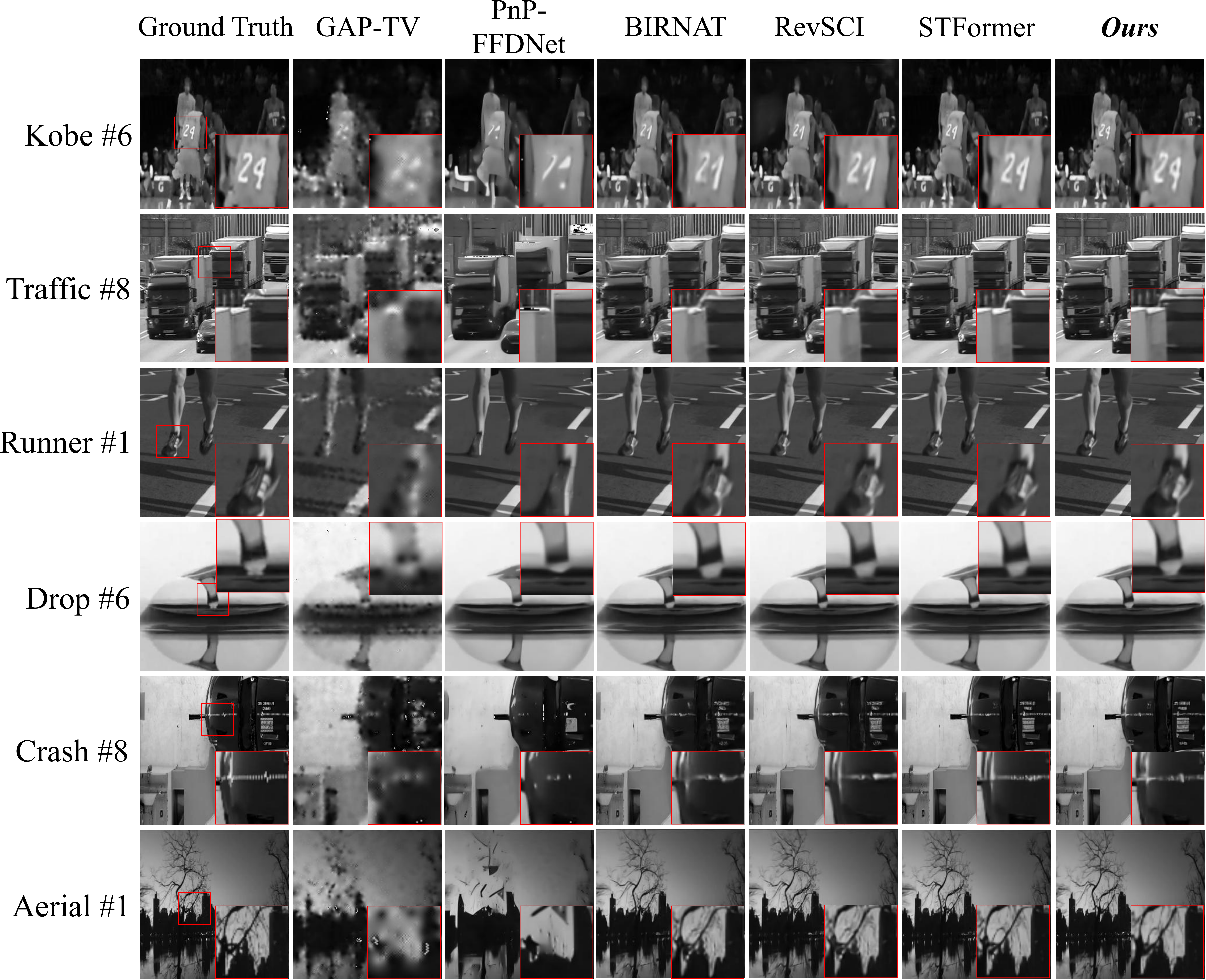}
    \caption{Reconstructed frames of the simulated USS measurements using different methods when Cr is 8. For a better view, we zoom in on a local area as shown in the small red boxes of each ground truth image, and do not show the small red boxes again for simplicity.}
    \label{fig:RS_USS_sim_cr10}
\end{figure*}

\subsection{Results on Simulated Data}
We present the reconstruction quality of different video SCI reconstruction methods, including model-based methods (GAP-TV~\cite{Yuan16ICIP_GAP}, PnP-FFDNet~\cite{yuan2020plug}, and PnP-FastDVDnet~\cite{yuan2021plug}) and deep learning-based algorithms (BIRNAT~\cite{cheng2020birnat}, RevSCI~\cite{cheng2021memory}, STFormer~\cite{wang2022spatial}, EfficientSCI~\cite{wang2023efficientsci}, and our proposed BSTFormer) on six simulated testing datasets when Cr is 8. Reconstruction quality and visualization results are summarized in Tab.~\ref{Tab:sim6} and Fig.~\ref{fig:RS_USS_sim_cr10}. For a fair comparison, we uniformly measure the inference time of all deep learning-based methods on the same NVIDIA RTX 3090 GPU.

We can make the following observations:
\begin{itemize}
\item[1)] As previously discussed, due to the faster inference time of the deep neural network, all deep learning-based methods have a much shorter inference time than the optimization-based ones. Even the fastest optimization-based method, PnP-FFDNet, requires 3.0s. However, its reconstruction quality is poor with  24.28dB in PSNR value which is 9.95dB lower than our BSTFormer. 
\item[2)] Although BIRNAT and RevSCI demonstrate impressive inference speeds, their reconstruction quality is unsatisfactory, with an average PSNR value below 31dB—more than 3~dB lower than BSTFormer.  
\item[3)] BSTFormer outperforms all previous methods, achieving PSNR values 1.02dB and 0.92dB higher than the previous SOTA methods, STFormer and EfficientSCI, respectively. Additionally, BSTFormer reduces inference time by approximately 55\% compared to STFormer and 29\% compared to EfficientSCI, making it a highly efficient network design.
\item[4)] For visualization purposes, we display  reconstructed grayscale video frames in Fig.~\ref{fig:RS_USS_sim_cr10}, we can see from the zooming areas in each selected video frame that BSTFormer provides much clearer reconstructed images than previous SOTA methods. Especially for the \texttt{Traffic}, \texttt{Crash}, and \texttt{Aerial} scenes, we can easily observe sharper edges and more details than other methods. 
\item[5)] In addition to USS data, we evaluate our BSTFormer on RS data. As shown in 
Tab.~\ref{Tab:para_floats_RS_main}, $i)$ Our BSTFormer marginally outperforms BIRNAT, RevSCI, EfficientSCI-S, and STFormer-S by 2.07dB, 1.46dB, 1.16dB, and 1.44dB, respectively. $ii)$ From the computational complexity and PSNR comparison, we infer that our BSTFormer can act as a ``medium-sized model'' between STFormer-S and STFormer-B, as well as EfficientSCI-T and EfficientSCI-B. $iii)$ Compared with EfficientSCI-S, our BSTFormer can achieve comparable reconstruction quality with similar computational complexity. In conclusion, our proposed BSTFormer \textit{achieves competitive reconstruction quality} on RS data.
\end{itemize}

\begin{table}[!ht]
  \setlength\tabcolsep{1.5pt}
  \caption{{Computational complexity and reconstruction quality of several reconstruction algorithms when testing on the simulated RS data.}}
  \centering
  {
  \centering
  \begin{tabular}{c|cc|ccc}
  \toprule
  Method &PSNR &SSIM & Params(M) &FLOPs(G) &Test time(s)
  \\
  \midrule
  BIRNAT~\cite{cheng2020birnat}  &33.31 &0.951 &4.13 &390.56 &0.10
  \\
  RevSCI~\cite{cheng2021memory}  &33.92 &0.956 &5.66 &766.95 &0.19
  \\
  \midrule
  STFormer-S~\cite{wang2022spatial}  &33.94 &0.958 & 1.22 &193.47 &0.14
  \\
  STFormer-B~\cite{wang2022spatial}  &36.34 &0.974 & 19.48 &3060.75 &0.49
  \\
  \midrule
  EfficientSCI-T~\cite{wang2023efficientsci} &34.22 &0.961 & 0.95 &142.18 &0.07
  \\
  EfficientSCI-S~\cite{wang2023efficientsci} &35.51 &0.970 & 3.78 &563.87 &0.15
  \\
  EfficientSCI-B~\cite{wang2023efficientsci}  &36.48 &0.975 &8.82 &1426.38 &0.31
  \\
  \midrule
  \rowcolor{lightgray}
  Our BSTFormer  &35.38 &0.969 &4.25 &710.87 &0.22
  \\
  \bottomrule
  \end{tabular}
  }
  \label{Tab:para_floats_RS_main}
\end{table}

\begin{figure*}[t]
    \centering 
    \includegraphics[width=.75\linewidth]{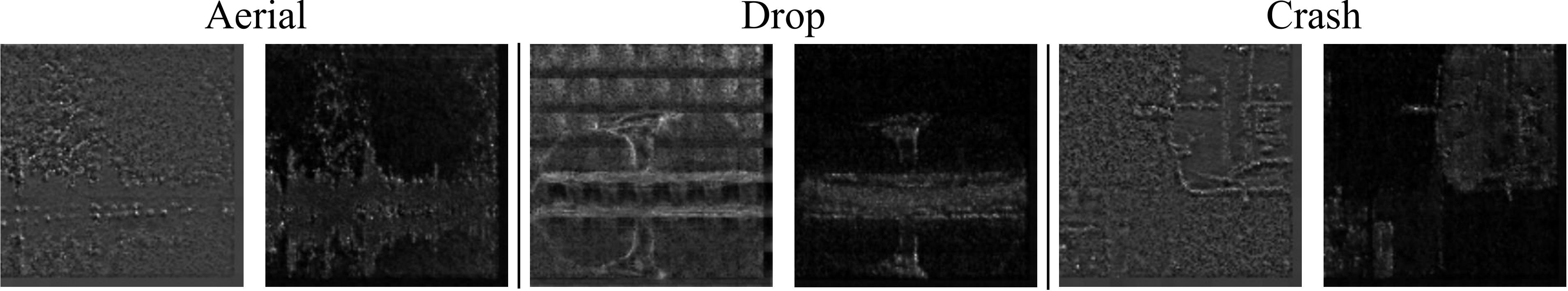}
    \caption{{The output feature maps of the LBA (left entry of each scene) and GSA (right entry of each scene) branches when testing on the simulated USS data. The LBA branch effectively extracts local features uniformly across the entire scene. The GSA branch primarily focuses on features at specific positions, such as object edges.} 
    }
  \label{fig:attention_block_main}
\end{figure*}

\begin{table}[!ht]
\setlength\tabcolsep{2pt}
  \caption{{Ablation study on our BSTFormer, where reconstruction quality on the simulated USS data and computational cost are used as evaluation metrics.}}
  \centering
  {
  \centering
  \begin{tabular}{ccc|cc|ccc}
  \toprule
  LBA
  & GSA
  & GTA
  & PSNR 
  & SSIM
  & Params(M) 
  & FLOPs(G)
  & Test time(s)
  \\
  \midrule
   &\checkmark &\checkmark  
    &33.29 &0.961 &5.83 &932.79  &0.23
  \\
  \checkmark & &\checkmark 
     &33.38 & 0.962 &5.83 &932.79 &0.23
  \\
  \checkmark &\checkmark & 
     &33.40 &0.962 &5.83 &933.35 &0.23
  \\
  \checkmark &\checkmark &\checkmark
     &34.23 &0.965 &4.25 &710.87 &0.22
  \\
  \bottomrule
  \end{tabular}
  }
  \label{Tab:ablation}
\end{table}

\begin{table}[!htbp]
  \setlength\tabcolsep{3pt}
  \renewcommand{\arraystretch}{1.5}
  \caption{The PSNR improvement in dB after BSTFormer is performed on the simulated USS data.}
  \centering
  {
  \centering
  \begin{tabular}{c|cccccc|c}
  \hline
  Dataset 
  & Kobe 
  & Traffic 
  & Runner 
  & Drop 
  & Crash 
  & Aerial 
  & Average 
  \\
  \hline
  Our BSTFormer
  & 20.70
  & 23.12
  & 29.58
  & 41.18
  & 23.18
  & 20.94
  & 26.45
  \\
  \hline 
  \end{tabular}
  }
  \label{Tab:xe_xout}
\end{table}

\subsection{Ablation Study}
In this section, we conduct ablation experiments on the proposed BSTFormer (including LBA, GSA, and GTA) to verify the effectiveness of our network design. Reconstruction quality and computational cost on the simulated testing datasets are summarized in Tab.~\ref{Tab:ablation}. Additionally, we provide some intermediate results from the LBA and GSA branches in Fig.~\ref{fig:attention_block_main}. Finally, we replace our BSTFormer with SOTA image restoration models (including FocalNet~\cite{cui2023focal}, GRL~\cite{li2023efficient}, and CODEFormer~\cite{zhao2023comprehensive}), and keep initialization process constant. Note that, to fairly compare with our proposed BSTFormer, we adapt the image restoration models by modifying the input from 3-channel image to 8-channel video frames. We then re-train these adapted models using the \texttt{DAVIS2017} dataset. The following observations can be made:
\begin{itemize}
\item[1)] Removing the LBA module, which is used to extract local spatial features, reduces the PSNR value by 0.94~dB. This result demonstrates that local spatial features are crucial for achieving better reconstruction quality.
\item[2)] Removing the GSA module, which establishes global spatial correlations, leads to a PSNR reduction of 0.85~dB. This highlights the importance of GSA for exploring sparsity within USS measurements.
\item[3)] The GTA module, responsible for building global temporal correlations, is also essential. When removed, the PSNR value drops by approximately 0.83~dB, further verifying its significance.
\item[4)] Interestingly, removing the Transformer branches (LBA, GSA, or GTA) increases the computational cost (FLOPs) by about 31$\%$. The underlying reason is that splitting the input features into three parts results in greater computational cost reduction compared to splitting into two parts.
\item[5)] From Fig.~\ref{fig:attention_block_main}, we learn that: $i)$ The LBA branch effectively extracts local features uniformly across the entire scene. $ii)$ The GSA branch primarily focuses on features at specific positions, such as object edges. $iii)$ The output feature maps from the LBA and GSA branches are complementary, enabling the extraction of features at multiple granularities, which is beneficial for reconstruction.
\item[6)] As shown in Tab.~\ref{Tab:xe_xout}, the average gain between $\Xmat_e$ and $\Xmat_{out}$ after applying the proposed BSTFormer is 26.45dB when tested on the simulated testing datasets. This result verifies the effectiveness of our BSTFormer.
\item[7)] As shown in Tab.~\ref{Tab:image_restoration_main}, our BSTFormer outperforms FocalNet~\cite{cui2023focal}, GRL~\cite{li2023efficient}, and CODEFormer~\cite{zhao2023comprehensive} by 3.02dB, 0.65dB, and 1.5dB, demonstrating the superior performance of our proposed BSTFormer.
\end{itemize}

\begin{table*}[t]
  \renewcommand{\arraystretch}{1.0}
  \caption{The average PSNR in dB (left entry) and SSIM (right entry) of different image restoration methods when testing on the simulated USS data.}
  \centering
  \resizebox{\textwidth}{!}
  {
  \centering
  \begin{tabular}{c|c|c|c|c|c|c|c}
  \hline
  Dataset 
  & Kobe 
  & Traffic 
  & Runner 
  & Drop 
  & Crash 
  & Aerial 
  & Average 
  \\
  \hline
  FocalNet~\cite{cui2023focal} (ICCV 2023)
  & 29.19, 0.910
  & 26.14, 0.984
  & 35.91, 0.964
  & 40.09, 0.980 
  & 27.52, 0.920 
  & 28.43, 0.901 
  & 31.21, 0.929
  \\
  \hline 
  GRL~\cite{li2023efficient} (CVPR 2023)
  & 30.89, 0.945  
  & 28.29, 0.937
  & 39.12, 0.984
  & 43.98, 0.994
  & 29.05, 0.954
  & 30.12, 0.938
  & 33.58, 0.959
  \\
  \hline
  CODEFormer~\cite{zhao2023comprehensive} (CVPR 2023) 
  & 30.59, 0.941  
  & 27.64, 0.927
  & 38.09, 0.979
  & 42.49, 0.991
  & 28.30, 0.944
  & 29.23, 0.923
  & 32.73, 0.951
  \\
  \hline
  \rowcolor{lightgray}
  Our BSTFormer
  & {32.61}, {0.962}
  & {29.82}, {0.955}
  & {39.77}, {0.984}
  & {43.94}, {0.993}
  & {29.20}, {0.955}
  & {30.04}, {0.939}
  & {34.23}, {0.965}
  \\
  \hline 
  \end{tabular}
  }
  \label{Tab:image_restoration_main}
\end{table*}

\subsection{Next Step of Video SCI}
Extensive experiments have demonstrated the significant advantages of video SCI for capturing high-speed videos using a low-speed camera. Thanks to deep learning, especially for the powerful Transformer networks, high-quality and stable reconstruction can be achieved when Cr is lower than 30, and good results can be obtained when Cr is reaching 50. Additionally, under strong lighting conditions, the USS strategy provides high-quality reconstruction results when Cr is lower than 30.
\begin{itemize}
    \item From the hardware perspective, the price we paid in video SCI is the high-speed modulation, implemented by DMD in our paper but can also be conducted by a shifting mask~\cite{llull2013coded}. 
    \item Looking forward, the underlying principle of I2P is to readout a small fraction of pixels at each high-speed frame by keeping the data throughput small. This can be achieved by the new generation of global shutter CMOS camera~\cite{seo20222}.
    Therefore, on-chip sensor design will be a good choice for the next generation vision cameras.
    \item Imaging is a tool to capture scenes, while the ultimate goal is to finish a task. Therefore, high quality reconstruction is one goal but not necessary in some task-driven SCI designs~\cite{zhang2022compressive}.
\end{itemize}

\section{Concluding Remarks}
\label{Sec:cond}
Over the past decades, advances in computational imaging have enabled snapshot imaging systems to capture high-speed videos~\cite{chen2022physics,qiao2020deep,wang2024situ,luo2024snapshot}, hyperspectral images~\cite{wang2022snapshot,meng2020end}, and holography~\cite{dou2023coded,wittwer2022phase}. With deep learning, high-quality reconstruction results have been achieved using neural networks~\cite{wang2022spatial,wang2023efficientsci}, particularly with emerging Transformers, as demonstrated in this work for video compressive sensing.

SCI represents a physical layer process that maximizes information capacity per sample while minimizing system size, power consumption, and cost. A critical next step is to integrate SCI onto a chip, paving the way for next-generation machine vision systems. Towards this goal, we introduced USS masks into video SCI and showed that, under strong lighting conditions and low Cr values, USS combined with our sparse Transformer-based reconstruction network delivers excellent results. However, light efficiency remains a challenge that requires further investigation.
Depending on specific applications~\cite{martel2020neural}, video SCI systems can utilize either random masks or ultra-sparse masks. For both patterns, moving beyond random designs towards optimized mask patterns is a promising direction.

Looking ahead, computational imaging, beyond SCI, will benefit from new sampling techniques for high-dimensional data. Another important direction is deploying video SCI reconstruction algorithms on resource-constrained devices, such as smartphones and autonomous vehicles. This necessitates the exploration of model compression techniques, including network pruning~\cite{wang2021neural,fang2023depgraph}, knowledge distillation~\cite{wang2022makes,rao2023parameter}, and low-bit quantization~\cite{li2021residual,wang2022deep}. 


\bibliographystyle{IEEEtran}
\bibliography{refs}

\end{document}